\documentclass[11pt,twocolumn]{article}

\usepackage[T1]{fontenc}
\usepackage[utf8]{inputenc}
\usepackage{lmodern}
\usepackage{microtype}
\usepackage[
    letterpaper,
    margin=0.75in
]{geometry}

\usepackage{amsmath,amssymb,amsfonts}
\usepackage{amsthm}
\usepackage{mathtools}

\usepackage{xcolor}

\usepackage{graphicx}
\usepackage{subcaption}
\usepackage{booktabs}
\usepackage{multirow}
\usepackage{multicol}
\usepackage{array}
\usepackage{makecell}
\usepackage{tikz}
\usepackage{caption}
\usepackage{pifont}
\usepackage{rotating}
\usepackage{adjustbox}
\newcommand{\xmark}{\ding{55}}%
\newcommand{\rowht}{\rule{0pt}{15ex}}
\newcommand{\rowhtmore}{\rule{0pt}{20ex}}
\definecolor{airforceblue}{RGB}{93,138,168}
\definecolor{bluegray}{rgb}{0.4, 0.6, 0.8}

\usepackage{algorithm}
\usepackage{algpseudocode}

\usepackage[numbers]{natbib}

\usepackage{xurl}
\usepackage[hidelinks]{hyperref}
\usepackage[nameinlink]{cleveref}

\usepackage{enumitem}

\usepackage{siunitx}

\title{CAPTOR: a vision-language foundation model for Chest-Abdomen-Pelvis Oncologic CT}
\author{
Séléna Jean-Mactoux$^{1,2}$ \and
Sophie Loizillon$^{2}$ \and
Ghina Jardali$^{1,3}$ \and
Amandine Crombé$^{3,4,5}$ \and
Nathalie Lassau$^{1,3}$ \and
Alexandre Bône$^{2}$\and\\
\small
$^{1}$ Biomaps, UMR1281, INSERM, CNRS, CEA, Université Paris-Saclay, Villejuif, France\\
\small
$^{2}$ Guerbet Research, Villepinte, France\\
\small
$^{3}$ Department of Diagnostic Oncologic Imaging, Gustave Roussy Institute, Villejuif, France\\
\small
$^{4}$ Department of Radiology, Pellegrin University Hospital, Bordeaux, France\\
\small
$^{5}$ Bordeaux Institute of Oncology, BRIC U1312, INSERM, Team SARCOTARGET, Bordeaux, \\\small France
}
\date{}
\begin{document}

\maketitle

\begin{abstract}
\hspace{0.4cm}\textbf{Objectives}

To develop and evaluate CAPTOR, a deep-learning vision-language foundation model for chest-abdomen-pelvis (CAP) CT interpretation in an oncologic setting, designed to detect important non-oncologic conditions, and incorporating an anatomically structured preprocessing strategy to improve whole-examination representation learning.

\vspace{5mm}
\textbf{Materials and Methods}

CAPTOR (\textbf{C}hest-\textbf{A}bdomen-\textbf{P}elvis
founda\textbf{T}ion model for \textbf{O}ncologic \textbf{R}adiology) is
a contrastive vision-language model, trained using a pretraining
strategy for oncologic CAP CT (in single- or multiple-series
acquisitions), on a multicenter retrospective dataset from North
America, Europe and Asia, predominantly in an oncologic context. CAPTOR
was compared with five existing CT foundation models, as well as a
larger-scale variant (CAPTOR-100K), and an ablation without the proposed
anatomical preprocessing step (CAPTOR-NoPre). Models were evaluated on
disease detection, image retrieval, and survival prediction using
TumorImagingBench. Classification and retrieval were assessed on
held-out internal and public test sets. Confidence intervals and
pairwise comparisons were obtained using patient-level bootstrap
resampling and permutation tests.

\vspace{5mm}
\textbf{Results}

CAPTOR was trained on 25,512 CT-report pairs and evaluated on 10,794
studies. It achieved higher classification AUCs (Area Under the Receiver
Operating Characteristic Curve) than the five state-of-the-art
foundation models on CAP (0.767 vs. 0.695 for the next-best model;
p\textless0.001) and oncologic CT subsets (0.784 vs. 0.734;
p\textless0.001), with superior retrieval performance on both subsets.
CAPTOR ranked first on three of four oncologic survival benchmarks.
Despite being trained on substantially fewer examinations, CAPTOR
achieved comparable classification performance to CAPTOR-100K. Removing
the proposed preprocessing strategy significantly reduced classification
performance, including on the institutionally held-out test subset,
where CAPTOR achieved a macro-AUC of 0.737, compared with 0.691 for
CAPTOR-NoPre.

\vspace{5mm}
\textbf{Conclusions}

Anatomically structured preprocessing enabled a region-specific
vision-language architecture to process whole-examination CAP CT,
including multiple-series acquisitions, and contributed to improved performance across multiple downstream tasks. CAPTOR outperformed
state-of-the-art CT foundation models on CAP and oncologic subsets on
classification and retrieval, while achieving comparable performance to
a substantially larger-scale variant. These findings highlight the
potential of anatomical preprocessing for adapting vision-language
architectures to whole-examination CT interpretation.

We release our preprocessing, training and inference code at the following url : https://github.com/Guerbet-AI/CAPTOR and the model weights at the following url : https://huggingface.co/Guerbet-AI/CAPTOR
\end{abstract}

\section{Introduction}
With a projected 63\% increase in cancer incidence by
2045~\cite{global_cancer_observatory_estimated_2024}, pressure intensifies on radiological practice,
especially for Chest-Abdomen-Pelvis (CAP) CT interpretation, a key
modality for oncologic staging, metastatic assessment, and treatment
monitoring~\cite{hopper_body_2000,american_society_of_clinical_oncology_asco_nodate,zer_early_2025,glynne-jones_rectal_2017}. This further strains the radiology
workforce, with a radiologist shortage documented across Europe and the
United States~\cite{bruls_workload_2020,makary_are_nodate}, and elevates interpretation error
risk, particularly under fatigue~\cite{patel_radiologists_2020}. Scalable AI tools
supporting multiple downstream tasks from a single examination are
increasingly recognized by major radiology societies for workflow
augmentation~\cite{neri_what_2019}.

Recent advances on foundation models, specifically vision-language
models (VLMs) enable joint image-text representation learning from CT
images and radiology reports through contrastive pretraining, without
manual annotations. CLIP~\cite{radford_learning_2021} methods applied to medical
context, such as CT-CLIP~\cite{hamamci_generalist_2026} or
Merlin~\cite{blankemeier_merlin_2026}, have demonstrated the potential of this
approach across various diagnostic and prognostic tasks.

However, most CT foundation models are trained on anatomically specific
datasets (e.g. Chest for CT-CLIP or Abdomen-Pelvis for
Merlin~\cite{blankemeier_merlin_2026}), limiting their applicability to CAP
protocols. Others, such as SPECTRE~\cite{claessens_scaling_2025} and
COLIPRI~\cite{wald_comprehensive_2026}, both transformer-based multimodal
foundation models, achieve broader anatomical coverage through
heterogeneous training sets, but without oncologic-rich training data.
While oncologic-specific foundation models exist, they primarily target
lesion detection, characterization, or treatment response rather than
whole-image analysis~\cite{li_foundation_2026,xiang_visionlanguage_2025}. Yet, radiologists
interpreting CAP CT scans in oncologic patients must also identify a
wide spectrum of non-cancerous findings, ranging from treatment-related
complications to age- and comorbidity-related conditions, as well as
incidental findings requiring further evaluation, which can be
overlooked under high workload and during cancer-focused examinations. A
gap therefore exists between~narrowly focused oncologic models,
and~generalist foundation models lacking exposure to multimorbidity-rich
oncologic CAP imaging, with no evidence that oncology patients require
dedicated foundation models.

To address this, we developed CAPTOR
(\textbf{C}hest-\textbf{A}bdomen-\textbf{P}elvis founda\textbf{T}ion
model for \textbf{O}ncologic \textbf{R}adiology). CAPTOR is based on the
Merlin architecture and is trained with a region-centered preprocessing
strategy leveraging crops centered on 12 different anatomical regions.
This lets the model learn from whole CAP examinations, including
multi-series acquisitions.~The training set comprised 25,512
predominantly oncologic CAP CT with radiology reports, from diverse
scanners and protocols, reflecting routine clinical practice. Our
central hypothesis was that this preprocessing, rather than training
data alone, drives CAPTOR's performance on oncologic CAP CT.

We investigated this hypothesis through four complementary experiments:

\begin{enumerate}
\def\labelenumi{\arabic{enumi}.}
\item
  \textbf{Comparison with state-of-the-art CT foundation models:}~We
  compared CAPTOR with five state-of-the-art CT foundation models,
  including region-specific (CT-CLIP~\cite{hamamci_generalist_2026},
  Merlin~\cite{blankemeier_merlin_2026}) and generalist models
  (SPECTRE~\cite{claessens_scaling_2025}, COLIPRI~\cite{wald_comprehensive_2026},
  CT-FM~\cite{pai_vision_2025}), across three downstream tasks (medical
  condition classification, image retrieval and 2-year post-treatment
  survival prediction).
\item
  \textbf{Ablation of anatomical preprocessing:}~We compared CAPTOR with
  CAPTOR-NoPre, trained on the same dataset but without the proposed
  anatomical preprocessing strategy, to assess its contribution to model
  performance.
\item
  \textbf{Institutional generalization:}~We evaluated CAPTOR and the
  state-of-the-art models on the Institutionally Held-Out Test Subset
  (IHOTS), comprising examinations from institutions not represented in
  CAPTOR\textquotesingle s training data.
\item
  \textbf{Effect of training data scale and composition:}~We compared
  CAPTOR with CAPTOR-100K, a larger-scale variant pretrained on
  additional public CT datasets, to investigate whether increasing
  training data volume and changing its clinical composition affected
  downstream performance.
\end{enumerate}

\section{Materials and Methods}
\subsection{Datasets}
This retrospective study comprised three dataset categories:

\begin{enumerate}
\def\labelenumi{(\roman{enumi})}
\item
  an internal multicenter image-report dataset for vision-language
  pretraining, collected under data use agreements with 67 institutions
  (11 in training and testing, 56 in testing only) across the United
  States, Spain, France, Japan and Vietnam
\item
  public image-report datasets for downstream experiments :
  Merlin~\cite{blankemeier_merlin_2026}, CT-RATE~\cite{hamamci_generalist_2026},
  INSPECT~\cite{huang_inspect_2023}and BIMCV-R~\cite{chen_bimcv-r_2024}
\item
  public oncologic benchmark datasets for two-year post-treatment
  survival prediction, using TumorImagingBench~\cite{aerts_foundation_2025}:
  NSCLC-Radiomics~\cite{aerts_data_2019} and
  NSCLC-Radiogenomics~\cite{bakr_data_2017} on chest CT, and
  C4KC-KiTS~\cite{heller_c4kc_2019} and
  Colorectal-Liver-Metastases~\cite{simpson_preoperative_2023} on abdominal CT.
\end{enumerate}

All datasets were used under their respective use terms. For the
internal dataset, all data were fully anonymized before analysis,
removing all direct and indirect identifiers such that re-identification
of individual patients was not reasonably possible, and data handling
complied with regional data protection regulations, including HIPAA and
GDPR where relevant.

The internal dataset was curated for oncologic CT interpretation,
predominantly from tumor lesion examinations, plus incidental pancreatic
cancer detection data. Studies' indication (oncologic vs. non-oncologic)
was inferred from reports using Mistral Large 3~\cite{mistral_ai_introducing_nodate}
(Mistral AI, Paris, France). The original internal dataset was randomly
split at the patient level into training (89\%) and test (11\%) sets. To
evaluate generalizability to data from unseen institutions, we increase
test set size with data from 56 additional institutions across the
United States, none of which contributed to the training set. We name
this cohort IHOTS, for Institutionally Held-Out Test Subset.

Appendices \ref{app:methods_internal_dataset}-\ref{app:methods_oncologic_dataset} provide details on all datasets

\subsection{CAPTOR Architecture}
We used the Merlin~\cite{blankemeier_merlin_2026} architecture, a contrastive VLM
using Clinical Longformer~\cite{li_clinical-longformer_2022} as text encoder, and an ImageNet-pretrained inflated ResNet152~\cite{carreira_quo_2017} as 3D image
encoder.

\subsection{CAPTOR Preprocessing}

\begin{figure*}[!tb]
    \centering
    \includegraphics[width=\linewidth]{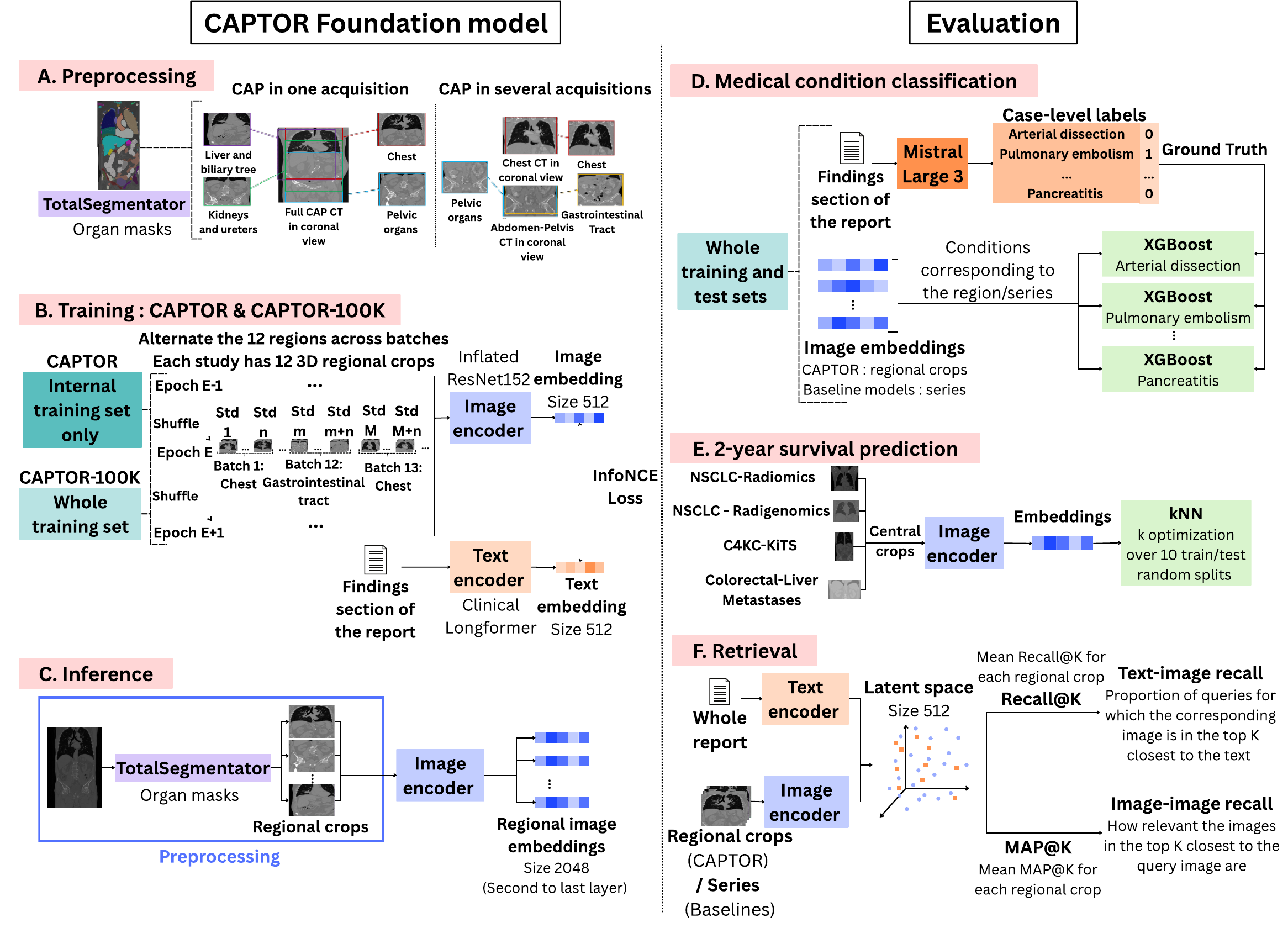}
    \caption{Overview of the CAPTOR pipeline. CT volumes are cropped into 12 anatomical sub-regions using TotalSegmentator~\cite{wasserthal_totalsegmentator_2023} masks (A); regional crops are paired with the Findings section of the corresponding report and CAPTOR is trained with a rotating-region contrastive objective (B). At inference, TotalSegmentator~\cite{wasserthal_totalsegmentator_2023}  masks are used to generate regional crops, and each CT yields one embedding per region (C). The resulting frozen image and text encoders are evaluated via XGBoost classification, k-nearest neighbor (kNN) survival prediction, and cross-modal retrieval (D, E, F). CAP = Chest-Abdomen-Pelvis; MAP@K = mean average precision at K, with K the number of retrieved images; std = Study; n = batch size; m = 11*n+1, M = 12*n+1}
    \label{fig:graphical_abstract}
\end{figure*}

Figure \ref{fig:graphical_abstract} provides an overview of the CAPTOR pipeline, encompassing data preprocessing, model training and evaluation.

CAPTOR's key innovation is its image preprocessing strategy: rather than
using fixed central crops like Merlin~\cite{blankemeier_merlin_2026},
TotalSegmentator~\cite{wasserthal_totalsegmentator_2023} masks were used to compute crop
centers for 12 anatomical regions such as `chest' and `liver and biliary
tree'. These crop centers were then used to dynamically crop CT scans
during training, also allowing us to use matching series depending on
the anatomical region (the thoracic CT when looking at the `chest'
region, and the Abdomen-Pelvis (AP) CT of the same study when looking at
the `liver and biliary tree' region for example).

Further details on preprocessing, the anatomical regions list and
details on report preprocessing are in \ref{app:methods_preprocessing}.

\subsection{CAPTOR Training}
CAPTOR was trained on the internal dataset with the contrastive
Information Noise-Contrastive Estimation (InfoNCE) loss, pairing
region-centered crops with the reports' \emph{Findings} parts. Batches
comprised studies, with the anatomical region rotating each batch.
Shuffling between epochs ensured all crops of each study contribute to
training. This strategy let the model learn representations from CAP
examinations, including multi-series, without architectural changes
(hyperparameters in Appendix \ref{app:methods_training}). 10\%
of the training set was randomly selected as validation set and early
stopping was used with a patience of 10 epochs.

\subsection{Evaluation - Comparison with
state-of-the-art}\label{evaluation---comparison-with-state-of-the-art}
CAPTOR was compared with five state-of-the-art foundation models :
Merlin~\cite{blankemeier_merlin_2026}, CT-CLIP~\cite{hamamci_generalist_2026},
SPECTRE~\cite{claessens_scaling_2025}, COLIPRI~\cite{wald_comprehensive_2026} and
CT-FM~\cite{pai_vision_2025}. SPECTRE~\cite{claessens_scaling_2025},
COLIPRI~\cite{wald_comprehensive_2026} and CT-CLIP~\cite{hamamci_generalist_2026} were
pretrained on the training splits of CT-RATE~\cite{hamamci_generalist_2026} (and
INSPECT~\cite{huang_inspect_2023}/Merlin~\cite{blankemeier_merlin_2026} for
SPECTRE~\cite{claessens_scaling_2025}), whose test splits are included in our test
set. Splits were respected, so there is no patient-level overlap.
Pretraining overlap between baseline models and our test set is detailed
in Appendix \ref{app:methods_baselines}.

To make sure our comparison to state-of-the-art models is fair, we used
all available series for every evaluation, using each model's method to
get the embeddings from each series and using the embeddings from the
matching series for each condition (e.g. embeddings from the Chest
series for chest conditions and from the AP series for AP conditions).

\vspace{5mm}
\textbf{Medical condition
classification}\label{medical-condition-classification}

We selected 26 medical conditions relevant to oncologic CAP CT
interpretation out of an initial list of 178 broadly covering a wide
spectrum of chest, abdominal and pelvic conditions, by severity and
utility as detection targets. More specifically, each medical condition
was assigned a criticality score by a radiologist taking into
consideration the urgency and severity, from 1 to 5 with 5 being the
highest risk conditions. The conditions with a criticality score of 3 or
more were then selected for this study.

Reference labels were built from reports (in or translated in English,
see Appendix \ref{app:methods_preprocessing}) using Mistral Large
3~\cite{mistral_ai_introducing_nodate}, a scalable supervision method shown to achieve
high agreement with expert annotations~\cite{flanders_evolution_2025,dorfner_comparing_2024}. To verify
extraction quality, we compared performance against
GPT-5.4~\cite{openai_introducing_2026} and GPT-4o mini~\cite{openai_gpt-4o_nodate} (OpenAI,
San Francisco, CA) on CT-RATE's~\cite{hamamci_generalist_2026} test set (composed
of 1,304 studies) using original labels.

For each condition, an XGBoost classifier was trained on embeddings from
the relevant anatomical crop. Performance was assessed on subsets
stratified by anatomical coverage (Chest, AP or CAP) and indication
(oncologic vs. non-oncologic), using the unweighted mean of
per-condition area under the receiver operating characteristic curve
(macro-AUC), restricted to conditions relevant to each subset.

Additional details in Appendices \ref{app:methods_label_extraction}-\ref{app:methods_generalization}.

\vspace{5mm}
\textbf{Retrieval}

Zero-shot text-image retrieval was evaluated using~Recall@K, the
proportion of text queries whose matching CT was retrieved among the top
K images, with~K in \{1,5,10,50,100\}. Zero-shot image-image retrieval
was evaluated using Mean Average Precision at K
(MAP@K~\cite{manzhos_average_2026}), rewarding retrieval of images sharing at
least one positive condition label (full definitions in Appendix \ref{app:methods_retrieval}). Fold-increase over random was used
for both metrics, better accounting for subset and K difficulty than raw
scores. SPECTRE~\cite{claessens_scaling_2025} could not be evaluated (unavailable
projection-layer weights), and CT-FM~\cite{pai_vision_2025}, a vision-only
model, was excluded from text-image retrieval.

\vspace{5mm}
\textbf{Survival prediction
benchmark}\label{survival-prediction-benchmark}

CAPTOR was evaluated on four two-year survival prediction tasks from
TumorImagingBench~\cite{aerts_foundation_2025}. Following
TumorImagingBench\textquotesingle s method, k-nearest neighbor (kNN)
classifiers (k optimized in \{1-50\}) were fit per model and dataset
over 10 random splits.

\subsection{Evaluation - Ablation of anatomical
preprocessing}\label{evaluation---ablation-of-anatomical-preprocessing}

To understand the effect of preprocessing versus the effect of training
data, another model was trained, CAPTOR-NoPre, using the exact same
architecture, training parameters and data as CAPTOR, but without using
our anatomical crops for preprocessing: only central crops were used.

\subsection{Evaluation -
Generalizability}\label{evaluation---generalizability}

Generalizability to data from unseen institutions was assessed on the
IHOTS cohort, comprising examinations from 56 institutions not
represented in CAPTOR\textquotesingle s training data, and
generalizability to non-oncologic data was also evaluated on chest CT
datasets CT-RATE~\cite{hamamci_generalist_2026} and
RAD-ChestCT~\cite{draelos_rad-chestct_2020}. Both non-oncologic chest datasets'
original labels were used, extracted from the reports with a fine-tuned
RadBERT-RoBERTa-4m~\cite{yan_radbert_2022} for CT-RATE~\cite{hamamci_generalist_2026}
and with a rule-based method for RAD-ChestCT~\cite{draelos_rad-chestct_2020}.

\subsection{\texorpdfstring{Evaluation - Effect of training data scale
and composition
}{Evaluation - Effect of training data scale and composition }}\label{evaluation---effect-of-training-data-scale-and-composition}

To assess the effect of larger, less clinically aligned training data,
CAPTOR-100K was trained on the complete training set (internal and
public) and compared with CAPTOR (internal only) and baseline models.

To study training data volume's impact on classification performance,
classifiers were trained on nested subsets of the whole training set
(10\%, 25\%, 50\% and 75\%), generated by iterative multilabel
stratification.

\subsection{Statistical analysis}\label{statistical-analysis}

95\% confidence intervals were estimated with a patient-level bootstrap
test (2,000 resamples; 1,000 for retrieval). Classification
per-condition pairwise comparisons were performed using the DeLong test,
and all other comparisons were performed with a patient-level two-sided
permutation test (2,000 permutations). All p-values were corrected per
task, using the Benjamini-Hochberg false discovery rate procedure
(significance set at p\textless0.05).

Analyses were performed using Python 3.12.12; Torch 2.7.0; NumPy 1.26.4;
Pandas 2.2.3, and scikit-learn 1.8.0.

\subsection{Code availability}
We release our preprocessing, training and inference code at this url : https://github.com/Guerbet-AI/CAPTOR and the model weights at this url : https://huggingface.co/Guerbet-AI/CAPTOR

\section{Results}
\subsection{Dataset characteristics}
\begin{figure}[!htbp]
    \centering
    \includegraphics[width=\linewidth]{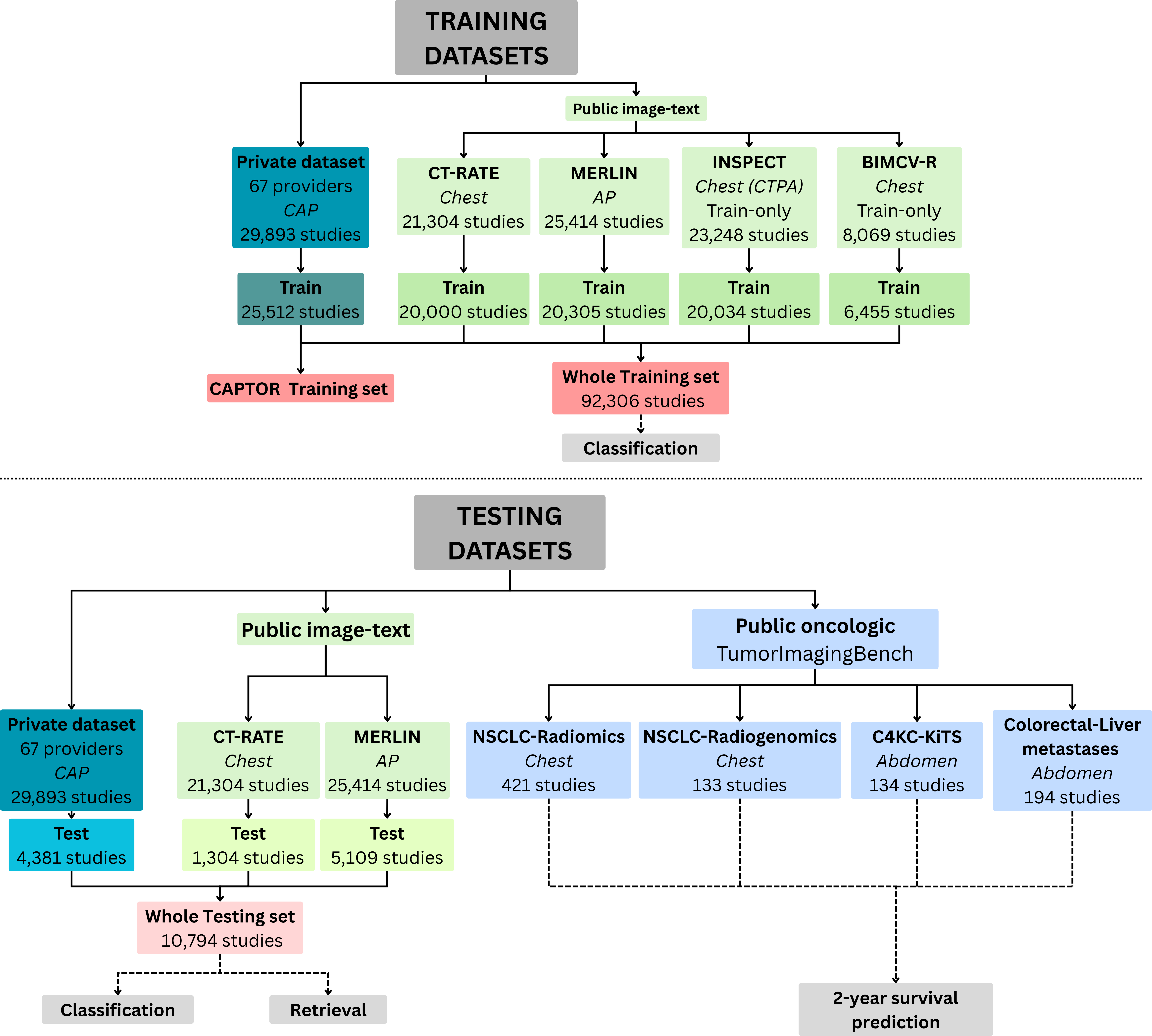}
    \caption{Flow chart showing the datasets used for training CAPTOR and for training/testing downstream tasks. CAP= Chest-Abdomen-Pelvis, AP= Abdomen-Pelvis}
    \label{fig:flow_chart}
\end{figure}

\begin{table*}[htbp]
\centering
\renewcommand{\arraystretch}{1.2}
\begin{adjustbox}{max width=\textwidth}
\begin{tabular}{ >{\centering\arraybackslash}m{1cm} >{\centering\arraybackslash}m{2.5cm} >{\centering\arraybackslash}m{1cm} >{\centering\arraybackslash}m{2.5cm} >{\centering\arraybackslash}m{3cm} >{\centering\arraybackslash}m{1.5cm} >{\centering\arraybackslash}m{3cm} >{\centering\arraybackslash}m{3.5cm} >{\centering\arraybackslash}m{3.5cm} >{\centering\arraybackslash}m{3.5cm} >{\centering\arraybackslash}m{2cm}}
\toprule
\multicolumn{2}{c}{\textbf{Dataset}}
 & \textbf{Subset}
  & \textbf{Studies}
  & \textbf{Age}
  & \textbf{Sex}
  & \textbf{Body Region}
  & \textbf{Indication}
  & \textbf{Contrast}
  & \textbf{Provenance}
  & \textbf{Report Section(s)}\\
\midrule
\multirow{3}{2cm}{%
  \raisebox{-3cm}{%
    \rotatebox[origin=c]{90}{Private}%
  }%
}
& \multirow[c]{3}{3cm}[-2.15cm]{\begin{tabular}[c]{@{}m{2cm}@{}}\centering Internal \\ dataset \end{tabular}} 
 & Total& \makecell[c]{29,893 studies \\ 32,689 series} & 66.0 [56.0-75.0] & F : 49\% \ M : 51\% &  
 \begin{tikzpicture}[x=0.03cm,y=0.35cm , baseline=0]
\fill[airforceblue] (0,6) rectangle (44.4,7);
\fill[airforceblue] (0,5) rectangle (23.7,6);
\fill[airforceblue] (0,4) rectangle (13.5,5);
\fill[airforceblue] (0,3) rectangle (8.8,4);
\fill[airforceblue] (0,2) rectangle (6.9,3);
\fill[airforceblue] (0,1) rectangle (1.8,2);
\fill[airforceblue] (0,0) rectangle (0.9,1);
\node[right] at (44.4,6.5) {44.4\%};
\node[right] at (23.7,5.5) {23.7\%};
\node[right] at (13.5,4.5) {13.5\%};
\node[right] at (8.8,3.5) {8.8\%};
\node[right] at (6.9,2.5) {6.9\%};
\node[right] at (1.8,1.5) {1.8\%};
\node[right] at (0.9,0.5) {0.9\%};
\node[left] at (0,6.5) {AP};
\node[left] at (0,5.5) {CAP};
\node[left] at (0,4.5) {Chest};
\node[left] at (0,3.5) {Unknown};
\node[left] at (0,2.5) {Abdomen};
\node[left] at (0,1.5) {CA};
\node[left] at (0,0.5) {Pelvis};
\end{tikzpicture}  & 70\% oncologic & Injected : 70\% \ Non-injected :  16\% \ Unknown :  14\% & \makecell[c]{Multi-centric \\ from 67 institutions in : \\ United States (66\%)\\Spain (16\%)\\France (14\%)\\Japan (3\%)\\Vietnam (1\%)} & \multirow[c]{3}{2cm}[-2.15cm]{Full report} \\ 
 & & Train & \makecell[c]{25{,}512 studies \\ 27{,}698 series}& 66.0 [56.0-75.0] & F : 49\% \ M :  51\% &  
\begin{tikzpicture}[x=0.03cm,y=0.35cm , baseline=0]
\fill[airforceblue] (0,6) rectangle (45.6,7);
\fill[airforceblue] (0,5) rectangle (21.9,6);
\fill[airforceblue] (0,4) rectangle (13.5,5);
\fill[airforceblue] (0,3) rectangle (8.8,4);
\fill[airforceblue] (0,2) rectangle (7.5,3);
\fill[airforceblue] (0,1) rectangle (1.8,2);
\fill[airforceblue] (0,0) rectangle (0.9,1);
\node[right] at (45.6,6.5) {45.6\%};
\node[right] at (21.9,5.5) {21.9\%};
\node[right] at (13.5,4.5) {13.5\%};
\node[right] at (8.8,3.5) {8.8\%};
\node[right] at (7.5,2.5) {7.5\%};
\node[right] at (1.8,1.5) {1.8\%};
\node[right] at (0.9,0.5) {0.9\%};
\node[left] at (0,6.5) {AP};
\node[left] at (0,5.5) {CAP};
\node[left] at (0,4.5) {Chest};
\node[left] at (0,3.5) {Unknown};
\node[left] at (0,2.5) {Abdomen};
\node[left] at (0,1.5) {CA};
\node[left] at (0,0.5) {Pelvis};
\end{tikzpicture}  & 69\% oncologic & Injected : 67\% \ Non-injected :  19\% \ Unknown :  13\% & \makecell[c]{Multi-centric \\ from 11 institutions} & \\ 
 & & Test& \makecell[c]{4{,}381 studies \\ 4,991 series} & 68.0 [57.0-77.0] & F : 48\% \ M :  52\% & 
 \begin{tikzpicture}[x=0.03cm,y=0.35cm , baseline=0]
\fill[airforceblue] (0,6) rectangle (38.0,7);
\fill[airforceblue] (0,5) rectangle (33.7,6);
\fill[airforceblue] (0,4) rectangle (13.8,5);
\fill[airforceblue] (0,3) rectangle (8.8,4);
\fill[airforceblue] (0,2) rectangle (3.1,3);
\fill[airforceblue] (0,1) rectangle (1.8,2);
\fill[airforceblue] (0,0) rectangle (0.8,1);
\node[right] at (38.0,6.5) {37.0\%};
\node[right] at (33.7,5.5) {34\%};
\node[right] at (13.8,4.5) {13.8\%};
\node[right] at (8.8,3.5) {8.8\%};
\node[right] at (3.1,2.5) {3.1\%};
\node[right] at (1.8,1.5) {1.8\%};
\node[right] at (0.8,0.5) {0.8\%};
\node[left] at (0,6.5) {AP};
\node[left] at (0,5.5) {CAP};
\node[left] at (0,4.5) {Chest};
\node[left] at (0,3.5) {Unknown};
\node[left] at (0,2.5) {Abdomen};
\node[left] at (0,1.5) {CA};
\node[left] at (0,0.5) {Pelvis};
\end{tikzpicture} & 75\% oncologic & Injected : 66\% \ Non-injected :  16\% \ Unknown :  17\% & \makecell[c]{Multi-centric \\ from 67 institutions \\ (56 additional \\ institutions unseen in \\ the training set)} & \\
\midrule

\multirow{3}{2cm}{%
  \raisebox{-8cm}{%
    \rotatebox[origin=c]{90}{Public}%
  }%
}
&\multirow[c]{3}{*}[-0.87cm]{CT-RATE}
  & Total  & \makecell[c]{21,304 studies\\ 25{,}432 series \\ 50{,}188 \ volumes} & 45.0 [35.0-62.0] & F : 42\% \ M :  58\% & \multirow[c]{3}{4cm}[-0.87cm]{\begin{tabular}[c]{@{}m{4cm}@{}}\centering Chest \end{tabular}} & \multirow[c]{3}{3cm}[-0.87cm]{\begin{tabular}[c]{@{}m{3cm}@{}}\centering Unknown\end{tabular}} & \multirow[c]{3}{4cm}[-0.87cm]{\begin{tabular}[c]{@{}m{4cm}@{}}\centering Non-injected\end{tabular}} & \multirow[c]{3}{4cm}[-0.87cm]{\begin{tabular}[c]{@{}m{4cm}@{}}\centering Mono-centric from Istanbul Medipol University Mega Hospital\end{tabular}} & \multirow[c]{3}{2cm}[-0.87cm]{ \begin{tabular}[c]{@{}m{2cm}@{}}\centering Findings \& Impressions\end{tabular} }\\
  & & Train  & \makecell[c]{20,000 studies\\ 24{,}128 series\\ 47{,}149 volumes} & 45.0 [34.0-61.0] & F : 42\% \ M :  58\% & & & & & \\
  & & Test & \makecell[c]{1,304 studies\\ 1,304 series \\ 3{,}039 volumes} & 45.0 [34.0-60.0] & F : 44\% \ M :  56\% &  &  & &  &\\
\cline{2-11}

& \multirow[c]{3}{*}[-0.55cm]{Merlin}
  & Total  & \makecell[c]{25,414 studies} & 55.0 [39.0-68.0] & F : 56\% \ M :  44\% & \multirow[c]{3}{4cm}[-0.55cm]{\begin{tabular}[c]{@{}m{4cm}@{}}\centering Abdomen \& Pelvis \end{tabular}} & \multirow[c]{3}{3cm}[-0.55cm]{\begin{tabular}[c]{@{}m{3cm}@{}}\centering Unknown\end{tabular}} & \multirow[c]{3}{4cm}[-0.55cm]{\begin{tabular}[c]{@{}m{4cm}@{}}\centering Injected\end{tabular}} & \multirow[c]{3}{4cm}[-0.55cm]{\begin{tabular}[c]{@{}m{4cm}@{}}\centering Multi-centric from Stanford Research Repository\end{tabular}} & \multirow[c]{3}{2cm}[-0.55cm]{ \begin{tabular}[c]{@{}m{2cm}@{}}\centering Findings \& Impressions\end{tabular} }\\
  & & Train  & \makecell[c]{20,305 studies} & 55.0 [39.0-68.0] & F : 56\% \ M :  44\% & & & & & \\
  & & Test & \makecell[c]{5,109 studies} & 55.0 [40.0-68.0] & F : 57\% \ M :  43\% &  &  & &  &\\
\cline{2-11}

& \multirow[c]{3}{3cm}[-1.45cm]{\begin{tabular}[c]{@{}m{2cm}@{}}\centering INSPECT \\ (CTPA) \end{tabular}}
  & Total  & \makecell[c]{23{,}248 studies} & 
  \begin{tikzpicture}[x=0.02cm,y=0.35cm , baseline=0]
\fill[airforceblue] (0,1) rectangle (13,2);
\fill[airforceblue] (0,2) rectangle (43,3);
\fill[airforceblue] (0,3) rectangle (25,4);
\node[left] at (0,1.5) {18-39};
\node[right] at (15.0,1.5) {13\%};
\node[left] at (0,2.5) {39-69};
\node[right] at (51.4,2.5) {43\%};
\node[left] at (0,3.5) {69-89};
\node[right] at (30.2,3.5) {25\%};
\end{tikzpicture} & F : 55\% \ M : 45\% & \multirow[c]{3}{4cm}[-1.45cm]{\begin{tabular}[c]{@{}m{4cm}@{}}\centering Chest \end{tabular}} & \multirow[c]{3}{3cm}[-1.45cm]{\begin{tabular}[c]{@{}m{3cm}@{}}\centering Pulmonary Embolism\end{tabular}} & \multirow[c]{3}{4cm}[-1.45cm]{\begin{tabular}[c]{@{}m{4cm}@{}}\centering Injected\end{tabular}} & \multirow[c]{3}{4cm}[-1.45cm]{\begin{tabular}[c]{@{}m{4cm}@{}}\centering Multi-centric from Stanford Research Repository\end{tabular}} & \multirow[c]{3}{2cm}[-1.45cm]{ \begin{tabular}[c]{@{}m{2cm}@{}}\centering Impressions only\end{tabular} }\\
  & & Train  & \makecell[c]{20,034 studies} & 
  \begin{tikzpicture}[x=0.02cm,y=0.35cm , baseline=0]
\fill[airforceblue] (0,1) rectangle (13,2);
\fill[airforceblue] (0,2) rectangle (43,3);
\fill[airforceblue] (0,3) rectangle (25,4);
\node[left] at (0,1.5) {18-39};
\node[right] at (15.0,1.5) {13\%};
\node[left] at (0,2.5) {39-69};
\node[right] at (51.4,2.5) {43\%};
\node[left] at (0,3.5) {69-89};
\node[right] at (30.2,3.5) {25\%};
\end{tikzpicture} & F 54\% \ M : 46\% & & & & & \\
  & & Test & \makecell[c]{3,214 studies} & 
  \begin{tikzpicture}[x=0.02cm,y=0.35cm , baseline=0]
\fill[airforceblue] (0,1) rectangle (14,2);
\fill[airforceblue] (0,2) rectangle (51,3);
\fill[airforceblue] (0,3) rectangle (32,4);
\node[left] at (0,1.5) {18-39};
\node[right] at (15.0,1.5) {14\%};
\node[left] at (0,2.5) {39-69};
\node[right] at (51.4,2.5) {51\%};
\node[left] at (0,3.5) {69-89};
\node[right] at (30.2,3.5) {32\%};
\end{tikzpicture} & F 56\% \ M : 44\% &  &  & &  &\\
\cline{2-11}

& BIMCV-R
  & Total 
  & 8{,}069
  & Unknown
  & Unknown
  & \begin{tabular}[c]{@{}m{4cm}@{}}\centering Chest \end{tabular} 
  & Unknown
  & Unknown
  & Multi-centric from Valencian Region Medical ImageBank 
  & Full report\\

\midrule
\midrule
\multirow{4}{2cm}{%
  \raisebox{-5.4cm}{%
    \rotatebox[origin=c]{90}{\makecell[c]{Public \\ TumorImagingBench}}%
  }%
}
& \rowht \multirow[c]{3}{*}[+1.6cm]{\makecell[c]{NSCLC - \\ Radiomics}} 
  & Total & \makecell[c]{421 studies} & 68.7 [61.2-75.9] & F : 31\% \ M :  69\% & \multirow[c]{3}{4cm}[+0.6cm]{\begin{tabular}[c]{@{}m{4cm}@{}}\centering Chest \end{tabular}} & \multirow[c]{3}{3cm}[+0.9cm]{\begin{tabular}[c]{@{}m{3.5cm}@{}}\centering NSCLC patients : \\- pre-radiation CT \\- GTV expert segmentation \end{tabular}} & \multirow[c]{3}{3.5cm}[+0.6cm]{\begin{tabular}[c]{@{}m{3.5cm}@{}}\centering Unknown\end{tabular}} & \multirow[c]{3}{3.5cm}[+0.6cm]{\begin{tabular}[c]{@{}m{3.5cm}@{}}\centering MAASTRO Clinic, The Netherlands\end{tabular}} & \multirow[c]{3}{2cm}[+0.6cm]{ \begin{tabular}[c]{@{}m{2cm}@{}}\centering No reports\end{tabular} }\\
\cline{2-11}

& \rowht \multirow[c]{3}{*}[+1.6cm]{\makecell[c]{NSCLC -\\ Radiogenomics}}
  &  Total  & \makecell[c]{133 studies} & 69.0 [64.0-76.0] & F : 25\% \ M :  75\% & \multirow[c]{3}{4cm}[+0.6cm]{\begin{tabular}[c]{@{}m{4cm}@{}}\centering Chest \end{tabular}} & \multirow[c]{3}{3cm}[+0.9cm]{\begin{tabular}[c]{@{}m{3.5cm}@{}}\centering  NSCLC patients : \\- pre-operative CT \\- GTV expert segmentation \end{tabular}} & \multirow[c]{3}{3.5cm}[+0.6cm]{\begin{tabular}[c]{@{}m{3.5cm}@{}}\centering Unknown\end{tabular}} & \multirow[c]{3}{3.5cm}[+0.9cm]{\begin{tabular}[c]{@{}m{3.5cm}@{}}\centering Stanford University School of Medicine and Palo Alto Veterans Affairs Healthcare System\end{tabular}} & \multirow[c]{3}{2cm}[+0.6cm]{ \begin{tabular}[c]{@{}m{2cm}@{}}\centering No reports\end{tabular} }\\
 \cline{2-11}

&\rowhtmore \multirow[c]{3}{*}[+2cm]{C4KC-KiTS}
  & Total  & \makecell[c]{134 studies} & 58.5 [50.0-68.0] & F : 41\% \ M :  59\% & \multirow[c]{3}{4cm}[+0.6cm]{\begin{tabular}[c]{@{}m{4cm}@{}}\centering Abdomen \end{tabular}} & \multirow[c]{3}{3cm}[+1.2cm]{\begin{tabular}[c]{@{}m{4cm}@{}}\centering Renal tumor patients : \\- partial/radical nephrectomy \\- pre-operative CT \\- tumor/kidney segmentations \end{tabular}} & \multirow[c]{3}{3.5cm}[+0.6cm]{\begin{tabular}[c]{@{}m{3.5cm}@{}}\centering Injected\end{tabular}} & \multirow[c]{3}{3.5cm}[+0.6cm]{\begin{tabular}[c]{@{}m{3.5cm}@{}}\centering University of Minnesota Medical Center\end{tabular}} & \multirow[c]{3}{2cm}[+0.6cm]{ \begin{tabular}[c]{@{}m{2cm}@{}}\centering No reports\end{tabular} }\\
 \cline{2-11}

&\rowhtmore \multirow[c]{3}{*}[+2cm]{\makecell[c]{Colorectal-Liver \\ Metastases}}
  & Total  & \makecell[c]{194 studies} & 61.0 [52.0-68.8] & F : 40\% \ M :  60\% & \multirow[c]{3}{4cm}[+0.6cm]{\begin{tabular}[c]{@{}m{4cm}@{}}\centering Abdomen \end{tabular}} & \multirow[c]{3}{3cm}[+1.2cm]{\begin{tabular}[c]{@{}m{4cm}@{}}\centering CRLM patients : \\-pre-operative CT \\- liver/tumor /vessels segmentation \\- CRLM resection \end{tabular}} & \multirow[c]{3}{3.5cm}[+0.6cm]{\begin{tabular}[c]{@{}m{3.5cm}@{}}\centering Injected\end{tabular}} & \multirow[c]{3}{3.5cm}[+0.6cm]{\begin{tabular}[c]{@{}m{3.5cm}@{}}\centering Memorial Sloan Kettering Cancer Center\end{tabular}} & \multirow[c]{3}{2cm}[+0.6cm]{ \begin{tabular}[c]{@{}m{2cm}@{}}\centering No reports\end{tabular} }\\

\bottomrule
\end{tabular}
\end{adjustbox}
\caption{Composition of the internal and public datasets used for training and testing. Percentages are reported at the study level. Train/test sets for TumorImagingBench are not included, as their original training/test splits were not used: survival prediction was instead assessed via k-nearest-neighbor classifiers across 10 random splits (Materials and Methods). The median age is shown, with inter-quartile range in brackets. CTPA = CT pulmonary angiogram; F=Female, M=Male, NSCLC = non-small cell lung cancer; GTV = gross tumor volume; CRLM = colorectal liver metastases.}
\label{Demographics}
\end{table*}

Table \ref{Demographics} summarizes dataset demographics and Figure \ref{fig:flow_chart} outlines how each dataset is used for pretraining and downstream tasks. The internal
dataset comprised 29,893 studies (32,689 series) from 67 institutions
across the United States (66\%), Spain (16\%), France (14\%), Japan
(3\%) and Vietnam (1\%), with a median age of 66.0 years, and was 49\%
female. Detailed characteristics by country of the internal dataset are
in Appendix \ref{app:results_internal_demographics}. The public
image-text datasets included CT-RATE~\cite{hamamci_generalist_2026} (21,304 chest
CT studies), Merlin~\cite{blankemeier_merlin_2026} (25,414 abdomen-pelvis CT
studies), INSPECT~\cite{huang_inspect_2023} (23,248 CT pulmonary angiograms),
and BIMCV-R~\cite{chen_bimcv-r_2024} (8,069 chest CT studies).

The resulting full training and testing sets including internal and
public datasets contained 92,306 studies and 10,794 studies
respectively. The testing set has a median age of 60 years
{[}Inter-Quartile Range 45-73{]} and was 52\% female. Among the testing
set, the IHOTS cohort represents 1,341 studies (31\% of the testing set
from our internal dataset) coming from 56 institutions not present in
the training set.

The four TumorImagingBench~\cite{aerts_foundation_2025} datasets comprised
between 133 and 421 studies each, covering lung
(NSCLC-Radiomics~\cite{aerts_data_2019}: 421 studies,
NSCLC-Radiogenomics~\cite{bakr_data_2017}: 133 studies) and abdominal
oncologic cohorts (C4KC-KiTS~\cite{heller_c4kc_2019}: 134 studies;
Colorectal-Liver Metastases~\cite{simpson_preoperative_2023}: 194 studies), all
without associated radiology reports.

\subsection{Comparison with
state-of-the-art}\label{comparison-with-state-of-the-art}

\textbf{Medical condition classification}
Performance assessment of label extraction on 1,304 reports from
CT-RATE's~\cite{hamamci_generalist_2026} test set, showed that
GPT-5.4~\cite{openai_introducing_2026}, Mistral Large 3~\cite{mistral_ai_introducing_nodate}, and
GPT-4o mini~\cite{openai_gpt-4o_nodate} achieved F1-scores of 0.917, 0.915, and
0.891 respectively (Appendix \ref{app:results_mistral_perfs}). Mistral Large 3~\cite{mistral_ai_introducing_nodate} was selected for its comparable
performance for approximately 13\% of the inference cost.

\begin{figure}[!t]
    \centering
    \includegraphics[width=\linewidth]{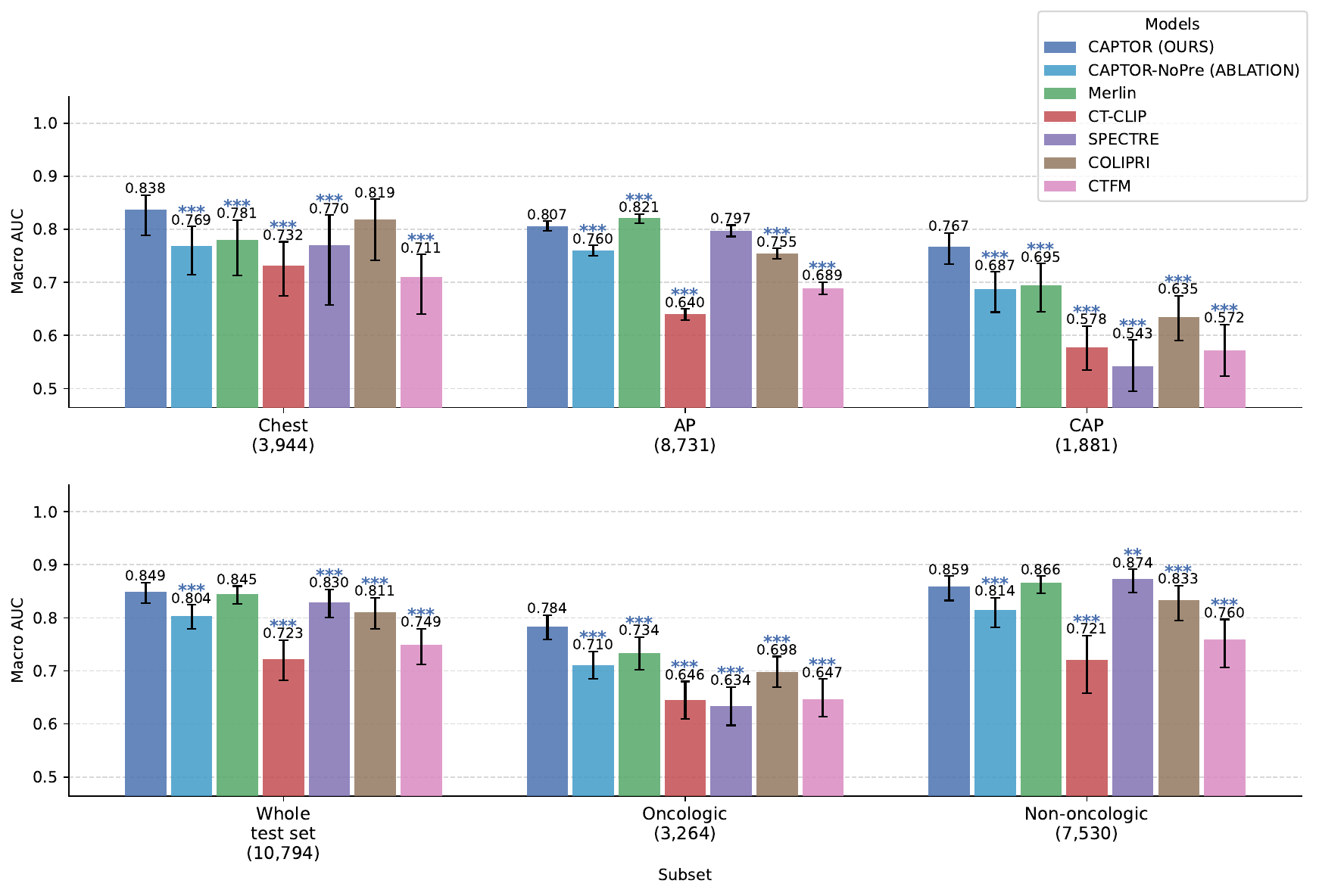}
    \caption{Bar plot showing the classification scores of CAPTOR and five state-of-the-art foundation models on our test set, stratified by visible anatomical region and clinical indication. The number of studies in each subset is in parentheses. One XGBoost was trained for each medical condition and model using the image embedding extracted with the model’s image encoder. For each subset and model, the macro-AUC is the unweighted mean of the AUC scores for each medical condition relevant to the subset among the 26 conditions studied. The confidence intervals and the pairwise comparison were obtained with a patient-level bootstrap and a patient-level permutation test respectively, both based on 2,000 iterations. AP = Abdomen-Pelvis, CAP=Chest-Abdomen-Pelvis.  *** = p < 0.001, ** = p < 0.01, * = p < 0.05}
    \label{fig:bar_plot_classif}
\end{figure}

As shown in Figure \ref{fig:bar_plot_classif}, CAPTOR achieved the highest overall performance,
reaching an AUC of 0.849 (95\% CI: 0.828-0.866) on the whole test set,
significantly (p\textless0.001) outperforming all models except for
Merlin~\cite{blankemeier_merlin_2026} (0.845, 95\% CI: 0.827-0.859).

Across anatomical regions, CAPTOR improved over
CT-CLIP~\cite{hamamci_generalist_2026} on the Chest subset (0.838, 95\% CI:
0.789-0.865 vs. 0.732, 95\% CI: 0.674 -- 0.777, p\textless0.001), while
Merlin~\cite{blankemeier_merlin_2026} outperformed CAPTOR on the AP subset (0.821,
95\% CI: 0.812-0.829 vs 0.807, 95\% CI: 0.797-0.816 for CAPTOR,
p\textless0.01). On the non-oncologic subset, CAPTOR was only
statistically significantly outperformed by SPECTRE~\cite{claessens_scaling_2025}
(0.859, 95\% CI: 0.833-0.879 vs. 0.874, 95\% CI: 0.848 -- 0.892,
p\textless0.01) despite its oncology-focused training.

Focusing specifically on our target use case, CAP images in an oncologic
context, CAPTOR statistically significantly outperformed all
state-of-the-art models, reaching an AUC of 0.767 (95\% CI: 0.734-0.793)
on the CAP subset and 0.784 (95\% CI: 0.759-0.805) on the oncologic
subset. Notably, Merlin~\cite{blankemeier_merlin_2026} outperformed the generalist
models (SPECTRE~\cite{claessens_scaling_2025}, COLIPRI~\cite{wald_comprehensive_2026} and
CT-FM~\cite{pai_vision_2025}) on both subsets despite being trained only on
AP images.

\begin{figure*}[t]
    \centering
    \includegraphics[width=0.7\linewidth]{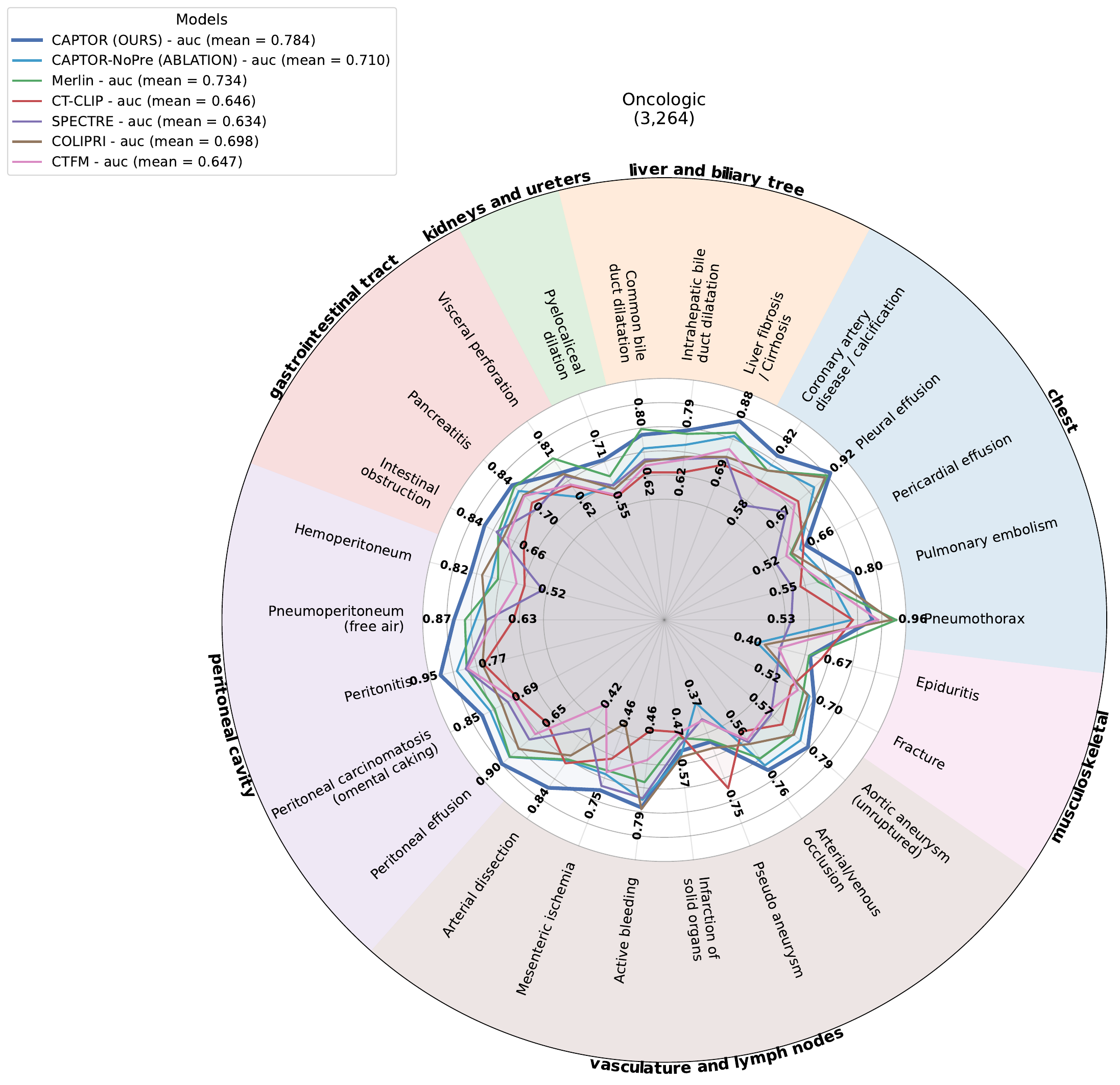}
    \caption{Per-condition classification performance of CAPTOR and five baselines on the oncologic subset, for each of the 26 medical conditions studied. One XGBoost was trained for each medical condition and model, using the appropriate image embedding (i.e. the corresponding regional crop for CAPTOR and the corresponding series for the baselines) extracted with the model’s image encoder.}
    \label{fig:radar_onco}
\end{figure*}

Figure \ref{fig:radar_onco} shows the details of per-condition classification performance
on the oncologic subset for CAPTOR and all baselines. Per-condition
proportions and significance counts are available in Appendices \ref{app:results_label_proportions}-\ref{app:results_stats_sig}.

\vspace{5mm}
\textbf{Retrieval}\label{retrieval}

CAPTOR achieved the highest text-image retrieval performance on CAP
(Recall@10 of 47.8, 95\% CI: 44.2--51.6), oncologic (83.4, 95\% CI:
78.4--88.6), and chest (64.3 , 95\% CI: 51.9--84.0) subsets, while
Merlin~\cite{blankemeier_merlin_2026} performed best on the whole test set (235,
95\% CI: 205--261), abdomen-pelvis (238, 95\% CI~: 230--245), and
non-oncologic subsets (236, 95\% CI: 199--272). Image-image retrieval
showed the same ranking, with CAPTOR achieving a MAP@10 of 1.30 (95\%
CI: 1.25--1.35) on the CAP subset, 1.41 (95\% CI: 1.37--1.46) on the
oncologic subset, and 1.23 (95\% CI: 1.20--1.27) on the chest subset
(Table \ref{Retrieval_table}).

\begin{table*}[htbp]
\renewcommand{\arraystretch}{1.3}
\centering
\begin{adjustbox}{max width=0.85\textwidth}
\begin{tabular}{cc|ccccc|cccccc}
\toprule
 & & \multicolumn{5}{c|}{Retrieval text-image} & \multicolumn{6}{c}{Retrieval image-image} \\
Dataset & K & \makecell[c]{CAPTOR \\ (OURS)} & \makecell[c]{CAPTOR - NoPre \\ (ABLATION)} & Merlin & CT-CLIP & COLIPRI & \makecell[c]{CAPTOR \\ (OURS)} & \makecell[c]{CAPTOR - NoPre \\ (ABLATION)} & Merlin & CT-CLIP & COLIPRI & CT-FM \\
\midrule
\multirow[c]{5}{*}{\centering \makecell{Whole test set \\ 10,794}} & 1 & \makecell{\underline{307} \\{\scriptsize (266--348)}} & \makecell{64.0 \\{\scriptsize (47.9--83.0)}} & \makecell{\textbf{722} \\{\scriptsize (622--817)}} & \makecell{20.0 \\{\scriptsize (8.88--31.1)}} & \makecell{188 \\{\scriptsize (104--281)}} & \makecell{\underline{1.79} \\{\scriptsize (1.73--1.83)}} & \makecell{1.72 \\{\scriptsize (1.64--1.74)}} & \makecell{\textbf{1.84} \\{\scriptsize (1.77--1.86)}} & \makecell{1.36 \\{\scriptsize (1.28--1.41)}} & \makecell{1.64 \\{\scriptsize (1.55--1.69)}} & \makecell{1.41 \\{\scriptsize (1.34--1.47)}} \\
 & 5 & \makecell{\underline{179} \\{\scriptsize (159--199)}} & \makecell{52.4 \\{\scriptsize (45.4--60.3)}} & \makecell{\textbf{353} \\{\scriptsize (307--394)}} & \makecell{12.2 \\{\scriptsize (6.60--18.7)}} & \makecell{98.4 \\{\scriptsize (57.8--144)}} & \makecell{\underline{1.43} \\{\scriptsize (1.38--1.44)}} & \makecell{1.38 \\{\scriptsize (1.32--1.38)}} & \makecell{\textbf{1.45} \\{\scriptsize (1.40--1.45)}} & \makecell{1.19 \\{\scriptsize (1.12--1.20)}} & \makecell{1.34 \\{\scriptsize (1.27--1.35)}} & \makecell{1.22 \\{\scriptsize (1.15--1.23)}} \\
 & 10 & \makecell{\underline{139} \\{\scriptsize (125--153)}} & \makecell{45.6 \\{\scriptsize (39.9--51.1)}} & \makecell{\textbf{235} \\{\scriptsize (205--261)}} & \makecell{11.3 \\{\scriptsize (6.71--16.6)}} & \makecell{70.1 \\{\scriptsize (42.6--101)}} & \makecell{\underline{1.47} \\{\scriptsize (1.45--1.51)}} & \makecell{1.41 \\{\scriptsize (1.38--1.44)}} & \makecell{\textbf{1.49} \\{\scriptsize (1.46--1.52)}} & \makecell{1.20 \\{\scriptsize (1.17--1.25)}} & \makecell{1.36 \\{\scriptsize (1.32--1.41)}} & \makecell{1.24 \\{\scriptsize (1.21--1.29)}} \\
 & 50 & \makecell{\underline{65.8} \\{\scriptsize (60.9--70.4)}} & \makecell{29.9 \\{\scriptsize (27.2--32.5)}} & \makecell{\textbf{77.1} \\{\scriptsize (67.8--84.8)}} & \makecell{8.02 \\{\scriptsize (4.97--11.8)}} & \makecell{30.7 \\{\scriptsize (21.7--40.5)}} & \makecell{\underline{1.60} \\{\scriptsize (1.57--1.65)}} & \makecell{1.52 \\{\scriptsize (1.49--1.57)}} & \makecell{\textbf{1.61} \\{\scriptsize (1.59--1.66)}} & \makecell{1.24 \\{\scriptsize (1.22--1.31)}} & \makecell{1.43 \\{\scriptsize (1.40--1.50)}} & \makecell{1.29 \\{\scriptsize (1.26--1.36)}} \\
 & 100 & \makecell{\underline{43.1} \\{\scriptsize (40.2--45.8)}} & \makecell{23.2 \\{\scriptsize (21.5--25.0)}} & \makecell{\textbf{45.9} \\{\scriptsize (40.9--50.2)}} & \makecell{6.62 \\{\scriptsize (4.24--9.54)}} & \makecell{21.0 \\{\scriptsize (16.0--26.5)}} & \makecell{\underline{1.61} \\{\scriptsize (1.59--1.66)}} & \makecell{1.53 \\{\scriptsize (1.50--1.58)}} & \makecell{\textbf{1.63} \\{\scriptsize (1.60--1.68)}} & \makecell{1.24 \\{\scriptsize (1.21--1.31)}} & \makecell{1.43 \\{\scriptsize (1.40--1.50)}} & \makecell{1.29 \\{\scriptsize (1.26--1.36)}} \\
\midrule
\multirow[c]{5}{*}{\centering \makecell{Chest \\ 3,944}} & 1 & \makecell{\underline{145} \\{\scriptsize (111--197)}} & \makecell{42.0 \\{\scriptsize (26.3--64.8)}} & \makecell{23.0 \\{\scriptsize (14.0--36.1)}} & \makecell{22.0 \\{\scriptsize (10.9--30.4)}} & \makecell{\textbf{184} \\{\scriptsize (111--231)}} & \makecell{\textbf{1.33} \\{\scriptsize (1.28--1.38)}} & \makecell{\underline{1.31} \\{\scriptsize (1.24--1.34)}} & \makecell{1.27 \\{\scriptsize (1.23--1.32)}} & \makecell{1.18 \\{\scriptsize (1.12--1.23)}} & \makecell{1.29 \\{\scriptsize (1.22--1.34)}} & \makecell{1.19 \\{\scriptsize (1.11--1.23)}} \\
 & 5 & \makecell{\underline{85.2} \\{\scriptsize (67.9--113)}} & \makecell{31.2 \\{\scriptsize (23.3--42.4)}} & \makecell{19.0 \\{\scriptsize (14.2--26.1)}} & \makecell{12.8 \\{\scriptsize (7.35--16.7)}} & \makecell{\textbf{92.0} \\{\scriptsize (58.4--114)}} & \makecell{\textbf{1.22} \\{\scriptsize (1.18--1.25)}} & \makecell{\underline{1.20} \\{\scriptsize (1.16--1.22)}} & \makecell{1.18 \\{\scriptsize (1.14--1.20)}} & \makecell{1.11 \\{\scriptsize (1.06--1.15)}} & \makecell{1.19 \\{\scriptsize (1.14--1.22)}} & \makecell{1.12 \\{\scriptsize (1.07--1.14)}} \\
 & 10 & \makecell{\textbf{64.3} \\{\scriptsize (51.9--84.0)}} & \makecell{26.9 \\{\scriptsize (20.5--36.0)}} & \makecell{15.3 \\{\scriptsize (11.9--20.0)}} & \makecell{11.6 \\{\scriptsize (7.04--15.0)}} & \makecell{\underline{63.3} \\{\scriptsize (41.1--78.0)}} & \makecell{\textbf{1.23} \\{\scriptsize (1.20--1.26)}} & \makecell{\underline{1.20} \\{\scriptsize (1.17--1.23)}} & \makecell{1.19 \\{\scriptsize (1.16--1.23)}} & \makecell{1.11 \\{\scriptsize (1.08--1.16)}} & \makecell{1.19 \\{\scriptsize (1.15--1.23)}} & \makecell{1.12 \\{\scriptsize (1.08--1.16)}} \\
 & 50 & \makecell{\textbf{28.6} \\{\scriptsize (24.4--35.4)}} & \makecell{16.2 \\{\scriptsize (12.9--20.6)}} & \makecell{9.10 \\{\scriptsize (7.79--11.1)}} & \makecell{8.08 \\{\scriptsize (5.50--9.92)}} & \makecell{\underline{25.7} \\{\scriptsize (18.8--30.0)}} & \makecell{\textbf{1.25} \\{\scriptsize (1.23--1.30)}} & \makecell{1.22 \\{\scriptsize (1.19--1.26)}} & \makecell{\underline{1.23} \\{\scriptsize (1.19--1.27)}} & \makecell{1.12 \\{\scriptsize (1.07--1.17)}} & \makecell{1.18 \\{\scriptsize (1.14--1.24)}} & \makecell{1.12 \\{\scriptsize (1.08--1.17)}} \\
 & 100 & \makecell{\textbf{18.8} \\{\scriptsize (16.2--22.9)}} & \makecell{12.5 \\{\scriptsize (10.2--15.7)}} & \makecell{6.71 \\{\scriptsize (5.86--7.90)}} & \makecell{6.58 \\{\scriptsize (4.65--7.93)}} & \makecell{\underline{16.8} \\{\scriptsize (13.3--19.1)}} & \makecell{\textbf{1.26} \\{\scriptsize (1.23--1.30)}} & \makecell{1.22 \\{\scriptsize (1.19--1.26)}} & \makecell{\underline{1.23} \\{\scriptsize (1.19--1.28)}} & \makecell{1.12 \\{\scriptsize (1.07--1.18)}} & \makecell{1.17 \\{\scriptsize (1.12--1.23)}} & \makecell{1.12 \\{\scriptsize (1.08--1.17)}} \\
\midrule
\multirow[c]{5}{*}{\centering \makecell{AP \\ 8,731}} & 1 & \makecell{\underline{270} \\{\scriptsize (241--305)}} & \makecell{53.0 \\{\scriptsize (39.8--67.0)}} & \makecell{\textbf{733} \\{\scriptsize (685--782)}} & \makecell{6.00 \\{\scriptsize (2.00--11.0)}} & \makecell{29.0 \\{\scriptsize (19.0--39.9)}} & \makecell{\underline{1.75} \\{\scriptsize (1.68--1.79)}} & \makecell{1.66 \\{\scriptsize (1.59--1.69)}} & \makecell{\textbf{1.81} \\{\scriptsize (1.73--1.83)}} & \makecell{1.26 \\{\scriptsize (1.20--1.30)}} & \makecell{1.55 \\{\scriptsize (1.48--1.58)}} & \makecell{1.30 \\{\scriptsize (1.24--1.35)}} \\
 & 5 & \makecell{\underline{157} \\{\scriptsize (148--168)}} & \makecell{44.8 \\{\scriptsize (39.0--50.5)}} & \makecell{\textbf{357} \\{\scriptsize (343--372)}} & \makecell{4.40 \\{\scriptsize (2.60--6.25)}} & \makecell{22.0 \\{\scriptsize (17.8--26.1)}} & \makecell{\underline{1.41} \\{\scriptsize (1.36--1.42)}} & \makecell{1.35 \\{\scriptsize (1.29--1.35)}} & \makecell{\textbf{1.43} \\{\scriptsize (1.38--1.44)}} & \makecell{1.14 \\{\scriptsize (1.07--1.13)}} & \makecell{1.29 \\{\scriptsize (1.23--1.29)}} & \makecell{1.17 \\{\scriptsize (1.10--1.17)}} \\
 & 10 & \makecell{\underline{122} \\{\scriptsize (116--129)}} & \makecell{38.8 \\{\scriptsize (34.9--42.8)}} & \makecell{\textbf{238} \\{\scriptsize (230--245)}} & \makecell{4.00 \\{\scriptsize (2.80--5.23)}} & \makecell{18.2 \\{\scriptsize (15.6--20.7)}} & \makecell{\underline{1.43} \\{\scriptsize (1.40--1.46)}} & \makecell{1.36 \\{\scriptsize (1.33--1.38)}} & \makecell{\textbf{1.45} \\{\scriptsize (1.42--1.48)}} & \makecell{1.13 \\{\scriptsize (1.11--1.17)}} & \makecell{1.29 \\{\scriptsize (1.26--1.32)}} & \makecell{1.17 \\{\scriptsize (1.14--1.20)}} \\
 & 50 & \makecell{\underline{56.6} \\{\scriptsize (54.8--58.2)}} & \makecell{25.4 \\{\scriptsize (24.1--26.6)}} & \makecell{\textbf{78.0} \\{\scriptsize (76.2--79.8)}} & \makecell{2.92 \\{\scriptsize (2.40--3.42)}} & \makecell{12.8 \\{\scriptsize (11.8--13.7)}} & \makecell{\underline{1.54} \\{\scriptsize (1.52--1.58)}} & \makecell{1.45 \\{\scriptsize (1.44--1.50)}} & \makecell{\textbf{1.56} \\{\scriptsize (1.54--1.60)}} & \makecell{1.16 \\{\scriptsize (1.15--1.21)}} & \makecell{1.35 \\{\scriptsize (1.34--1.40)}} & \makecell{1.20 \\{\scriptsize (1.19--1.25)}} \\
 & 100 & \makecell{\underline{36.9} \\{\scriptsize (36.0--37.8)}} & \makecell{19.6 \\{\scriptsize (18.8--20.3)}} & \makecell{\textbf{46.2} \\{\scriptsize (45.3--47.1)}} & \makecell{2.74 \\{\scriptsize (2.41--3.07)}} & \makecell{9.94 \\{\scriptsize (9.36--10.5)}} & \makecell{\underline{1.55} \\{\scriptsize (1.53--1.60)}} & \makecell{1.46 \\{\scriptsize (1.44--1.51)}} & \makecell{\textbf{1.57} \\{\scriptsize (1.55--1.62)}} & \makecell{1.16 \\{\scriptsize (1.15--1.21)}} & \makecell{1.35 \\{\scriptsize (1.33--1.40)}} & \makecell{1.19 \\{\scriptsize (1.18--1.24)}} \\
\midrule
\multirow[c]{5}{*}{\centering \makecell{CAP \\ 1,881}} & 1 & \makecell{\textbf{109} \\{\scriptsize (90.3--129)}} & \makecell{\underline{28.0} \\{\scriptsize (18.1--39.5)}} & \makecell{25.0 \\{\scriptsize (15.9--34.9)}} & \makecell{6.00 \\{\scriptsize (1.99--11.2)}} & \makecell{24.0 \\{\scriptsize (14.6--34.3)}} & \makecell{\textbf{1.48} \\{\scriptsize (1.38--1.54)}} & \makecell{\underline{1.44} \\{\scriptsize (1.34--1.50)}} & \makecell{1.36 \\{\scriptsize (1.29--1.45)}} & \makecell{1.23 \\{\scriptsize (1.13--1.30)}} & \makecell{1.35 \\{\scriptsize (1.25--1.41)}} & \makecell{1.21 \\{\scriptsize (1.12--1.29)}} \\
 & 5 & \makecell{\textbf{62.4} \\{\scriptsize (56.1--68.9)}} & \makecell{\underline{24.4} \\{\scriptsize (20.2--28.4)}} & \makecell{19.0 \\{\scriptsize (15.3--22.6)}} & \makecell{3.60 \\{\scriptsize (2.02--5.44)}} & \makecell{13.6 \\{\scriptsize (10.5--17.0)}} & \makecell{\textbf{1.29} \\{\scriptsize (1.22--1.32)}} & \makecell{\underline{1.26} \\{\scriptsize (1.19--1.29)}} & \makecell{1.23 \\{\scriptsize (1.16--1.26)}} & \makecell{1.13 \\{\scriptsize (1.06--1.17)}} & \makecell{1.21 \\{\scriptsize (1.14--1.24)}} & \makecell{1.13 \\{\scriptsize (1.06--1.17)}} \\
 & 10 & \makecell{\textbf{47.8} \\{\scriptsize (44.2--51.6)}} & \makecell{\underline{19.9} \\{\scriptsize (17.3--22.4)}} & \makecell{15.1 \\{\scriptsize (12.8--17.4)}} & \makecell{3.30 \\{\scriptsize (2.24--4.54)}} & \makecell{10.5 \\{\scriptsize (8.55--12.5)}} & \makecell{\textbf{1.30} \\{\scriptsize (1.25--1.35)}} & \makecell{\underline{1.27} \\{\scriptsize (1.22--1.32)}} & \makecell{1.24 \\{\scriptsize (1.19--1.29)}} & \makecell{1.13 \\{\scriptsize (1.08--1.19)}} & \makecell{1.20 \\{\scriptsize (1.16--1.26)}} & \makecell{1.13 \\{\scriptsize (1.08--1.19)}} \\
 & 50 & \makecell{\textbf{19.8} \\{\scriptsize (18.9--20.6)}} & \makecell{\underline{11.7} \\{\scriptsize (10.9--12.5)}} & \makecell{8.14 \\{\scriptsize (7.42--8.86)}} & \makecell{2.48 \\{\scriptsize (2.04--2.94)}} & \makecell{6.98 \\{\scriptsize (6.32--7.62)}} & \makecell{\textbf{1.34} \\{\scriptsize (1.30--1.41)}} & \makecell{\underline{1.30} \\{\scriptsize (1.26--1.37)}} & \makecell{1.28 \\{\scriptsize (1.24--1.35)}} & \makecell{1.13 \\{\scriptsize (1.09--1.20)}} & \makecell{1.20 \\{\scriptsize (1.16--1.27)}} & \makecell{1.14 \\{\scriptsize (1.10--1.21)}} \\
 & 100 & \makecell{\textbf{12.6} \\{\scriptsize (12.2--13.0)}} & \makecell{\underline{8.98} \\{\scriptsize (8.54--9.42)}} & \makecell{5.62 \\{\scriptsize (5.20--5.99)}} & \makecell{2.14 \\{\scriptsize (1.85--2.44)}} & \makecell{5.16 \\{\scriptsize (4.79--5.56)}} & \makecell{\textbf{1.33} \\{\scriptsize (1.28--1.39)}} & \makecell{\underline{1.29} \\{\scriptsize (1.24--1.36)}} & \makecell{1.27 \\{\scriptsize (1.23--1.34)}} & \makecell{1.12 \\{\scriptsize (1.07--1.19)}} & \makecell{1.18 \\{\scriptsize (1.13--1.25)}} & \makecell{1.13 \\{\scriptsize (1.08--1.20)}} \\
\midrule
\multirow[c]{5}{*}{\centering \makecell{Oncological \\ 3,264}} & 1 & \makecell{\textbf{224} \\{\scriptsize (195--253)}} & \makecell{\underline{53.0} \\{\scriptsize (39.8--66.9)}} & \makecell{38.0 \\{\scriptsize (26.1--51.2)}} & \makecell{6.00 \\{\scriptsize (1.99--11.0)}} & \makecell{41.0 \\{\scriptsize (29.1--53.9)}} & \makecell{\textbf{1.69} \\{\scriptsize (1.60--1.74)}} & \makecell{\underline{1.60} \\{\scriptsize (1.51--1.66)}} & \makecell{1.51 \\{\scriptsize (1.43--1.58)}} & \makecell{1.32 \\{\scriptsize (1.23--1.38)}} & \makecell{1.40 \\{\scriptsize (1.32--1.48)}} & \makecell{1.32 \\{\scriptsize (1.23--1.39)}} \\
 & 5 & \makecell{\textbf{115} \\{\scriptsize (106--125)}} & \makecell{\underline{44.6} \\{\scriptsize (38.8--50.2)}} & \makecell{27.6 \\{\scriptsize (23.1--32.2)}} & \makecell{6.20 \\{\scriptsize (4.19--8.43)}} & \makecell{22.8 \\{\scriptsize (18.7--27.1)}} & \makecell{\textbf{1.38} \\{\scriptsize (1.32--1.41)}} & \makecell{\underline{1.33} \\{\scriptsize (1.27--1.36)}} & \makecell{1.27 \\{\scriptsize (1.22--1.30)}} & \makecell{1.16 \\{\scriptsize (1.09--1.18)}} & \makecell{1.21 \\{\scriptsize (1.15--1.24)}} & \makecell{1.17 \\{\scriptsize (1.10--1.20)}} \\
 & 10 & \makecell{\textbf{83.4} \\{\scriptsize (78.4--88.6)}} & \makecell{\underline{38.3} \\{\scriptsize (34.7--42.0)}} & \makecell{21.5 \\{\scriptsize (18.8--24.2)}} & \makecell{4.60 \\{\scriptsize (3.32--5.90)}} & \makecell{16.9 \\{\scriptsize (14.4--19.5)}} & \makecell{\textbf{1.41} \\{\scriptsize (1.37--1.45)}} & \makecell{\underline{1.35} \\{\scriptsize (1.31--1.40)}} & \makecell{1.30 \\{\scriptsize (1.26--1.35)}} & \makecell{1.17 \\{\scriptsize (1.12--1.22)}} & \makecell{1.22 \\{\scriptsize (1.18--1.27)}} & \makecell{1.18 \\{\scriptsize (1.14--1.23)}} \\
 & 50 & \makecell{\textbf{33.5} \\{\scriptsize (32.4--34.6)}} & \makecell{\underline{21.8} \\{\scriptsize (20.8--22.9)}} & \makecell{11.2 \\{\scriptsize (10.3--12.0)}} & \makecell{3.28 \\{\scriptsize (2.79--3.78)}} & \makecell{9.94 \\{\scriptsize (9.16--10.7)}} & \makecell{\textbf{1.48} \\{\scriptsize (1.45--1.55)}} & \makecell{\underline{1.42} \\{\scriptsize (1.39--1.48)}} & \makecell{1.37 \\{\scriptsize (1.34--1.44)}} & \makecell{1.18 \\{\scriptsize (1.15--1.25)}} & \makecell{1.23 \\{\scriptsize (1.20--1.30)}} & \makecell{1.20 \\{\scriptsize (1.18--1.27)}} \\
 & 100 & \makecell{\textbf{21.0} \\{\scriptsize (20.5--21.5)}} & \makecell{\underline{15.7} \\{\scriptsize (15.1--16.3)}} & \makecell{8.09 \\{\scriptsize (7.61--8.60)}} & \makecell{2.90 \\{\scriptsize (2.56--3.23)}} & \makecell{7.38 \\{\scriptsize (6.93--7.88)}} & \makecell{\textbf{1.49} \\{\scriptsize (1.45--1.55)}} & \makecell{\underline{1.42} \\{\scriptsize (1.38--1.48)}} & \makecell{1.38 \\{\scriptsize (1.34--1.45)}} & \makecell{1.18 \\{\scriptsize (1.13--1.24)}} & \makecell{1.21 \\{\scriptsize (1.17--1.28)}} & \makecell{1.20 \\{\scriptsize (1.16--1.26)}} \\
\midrule
\multirow[c]{5}{*}{\centering \makecell{Non-oncologic \\ 7,530}} & 1 & \makecell{190 \\{\scriptsize (160--224)}} & \makecell{44.0 \\{\scriptsize (32.3--56.7)}} & \makecell{\textbf{745} \\{\scriptsize (619--880)}} & \makecell{22.0 \\{\scriptsize (7.78--36.1)}} & \makecell{\underline{191} \\{\scriptsize (93.7--288)}} & \makecell{\underline{1.69} \\{\scriptsize (1.61--1.74)}} & \makecell{1.60 \\{\scriptsize (1.51--1.64)}} & \makecell{\textbf{1.80} \\{\scriptsize (1.72--1.83)}} & \makecell{1.27 \\{\scriptsize (1.17--1.35)}} & \makecell{1.60 \\{\scriptsize (1.51--1.65)}} & \makecell{1.33 \\{\scriptsize (1.23--1.40)}} \\
 & 5 & \makecell{\underline{116} \\{\scriptsize (102--132)}} & \makecell{34.2 \\{\scriptsize (29.0--40.4)}} & \makecell{\textbf{365} \\{\scriptsize (308--423)}} & \makecell{14.4 \\{\scriptsize (6.30--22.3)}} & \makecell{99.0 \\{\scriptsize (53.6--142)}} & \makecell{\underline{1.37} \\{\scriptsize (1.31--1.38)}} & \makecell{1.30 \\{\scriptsize (1.24--1.32)}} & \makecell{\textbf{1.43} \\{\scriptsize (1.37--1.44)}} & \makecell{1.14 \\{\scriptsize (1.05--1.16)}} & \makecell{1.31 \\{\scriptsize (1.24--1.33)}} & \makecell{1.17 \\{\scriptsize (1.09--1.19)}} \\
 & 10 & \makecell{\underline{90.3} \\{\scriptsize (80.5--101)}} & \makecell{26.8 \\{\scriptsize (23.1--31.4)}} & \makecell{\textbf{236} \\{\scriptsize (199--272)}} & \makecell{11.9 \\{\scriptsize (5.55--18.2)}} & \makecell{69.3 \\{\scriptsize (38.8--98.1)}} & \makecell{\underline{1.42} \\{\scriptsize (1.38--1.46)}} & \makecell{1.35 \\{\scriptsize (1.30--1.39)}} & \makecell{\textbf{1.48} \\{\scriptsize (1.44--1.51)}} & \makecell{1.16 \\{\scriptsize (1.11--1.22)}} & \makecell{1.35 \\{\scriptsize (1.30--1.40)}} & \makecell{1.20 \\{\scriptsize (1.15--1.25)}} \\
 & 50 & \makecell{\underline{43.0} \\{\scriptsize (39.7--46.7)}} & \makecell{17.5 \\{\scriptsize (16.0--19.1)}} & \makecell{\textbf{74.2} \\{\scriptsize (64.0--84.8)}} & \makecell{8.24 \\{\scriptsize (4.10--12.4)}} & \makecell{27.6 \\{\scriptsize (18.7--36.5)}} & \makecell{\underline{1.52} \\{\scriptsize (1.48--1.57)}} & \makecell{1.43 \\{\scriptsize (1.39--1.49)}} & \makecell{\textbf{1.58} \\{\scriptsize (1.54--1.63)}} & \makecell{1.18 \\{\scriptsize (1.14--1.26)}} & \makecell{1.41 \\{\scriptsize (1.36--1.48)}} & \makecell{1.22 \\{\scriptsize (1.18--1.31)}} \\
 & 100 & \makecell{\underline{28.7} \\{\scriptsize (27.0--30.6)}} & \makecell{13.2 \\{\scriptsize (12.3--14.1)}} & \makecell{\textbf{42.8} \\{\scriptsize (37.2--48.5)}} & \makecell{6.41 \\{\scriptsize (3.27--9.48)}} & \makecell{18.7 \\{\scriptsize (13.6--23.5)}} & \makecell{\underline{1.53} \\{\scriptsize (1.49--1.59)}} & \makecell{1.44 \\{\scriptsize (1.40--1.51)}} & \makecell{\textbf{1.59} \\{\scriptsize (1.55--1.64)}} & \makecell{1.18 \\{\scriptsize (1.13--1.27)}} & \makecell{1.41 \\{\scriptsize (1.36--1.49)}} & \makecell{1.23 \\{\scriptsize (1.18--1.31)}} \\
\midrule
\end{tabular}
\end{adjustbox}
\caption{Zero-shot text-image (Recall@K) and image-image (MAP@K) retrieval performance of CAPTOR, CAPTOR-NoPre (i.e. model trained on the same dataset as CAPTOR but without our anatomical crops preprocessing) and four state-of-the-art models, expressed as fold-increase over the random baseline (best in bold and second best underlined) with 95\% confidence intervals in parentheses, by anatomical and oncologic subset. K is the number of retrieved images. SPECTRE~\cite{claessens_scaling_2025} could not be evaluated (unavailable projection-layer weights) and CT-FM~\cite{pai_vision_2025}, a vision model, was excluded from text-image retrieval.}
\label{Retrieval_table}
\end{table*}

\vspace{5mm}
\textbf{Survival prediction
benchmark}\label{survival-prediction-benchmark-1}

\begin{figure*}[!t]
    \centering
    \includegraphics[width=\linewidth]{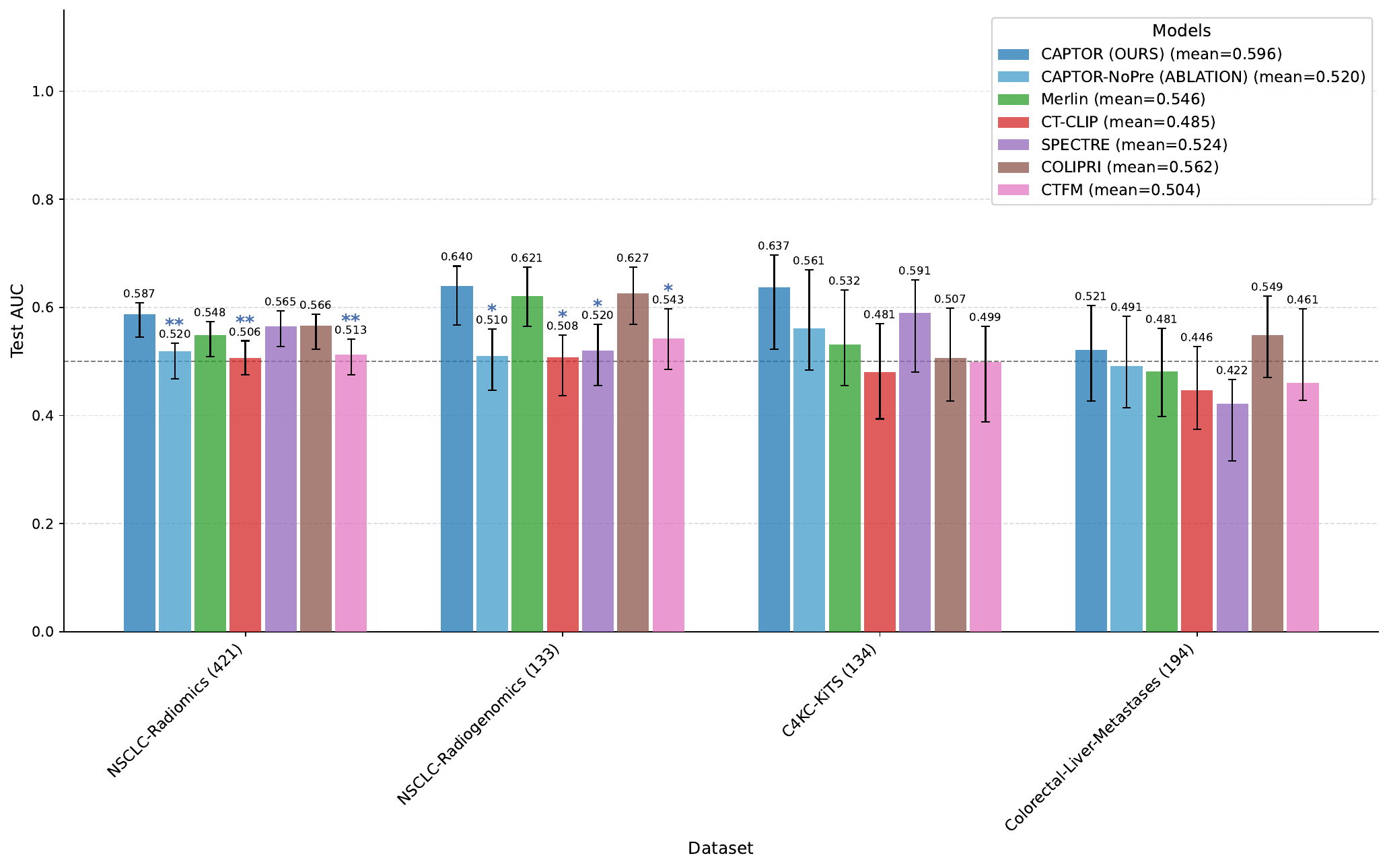}
    \caption{Two-year survival-prediction performance (AUC) of CAPTOR and five baseline models on four oncologic TumorImagingBench cohorts (NSCLC-Radiomics21, NSCLC-Radiogenomics22, C4KC-KiTS23, Colorectal-Liver-Metastases24), using kNN classifiers on frozen image embeddings, averaged over 10 random splits.  Statistical significance was assessed using a permutation test (2,000 permutations). The number of studies for each dataset is indicated between parentheses. ** = p < 0.01, * = p < 0.05}
    \label{fig:bar_plot_tumor}
\end{figure*}

CAPTOR ranked first on three of the four prognostic tasks, achieving
AUCs of~0.640~(95\% CI: 0.567--0.676) on
NSCLC-Radiogenomics~\cite{bakr_data_2017},~0.637~(95\% CI: 0.523--0.697)
on C4KC-KiTS, and~0.587~(95\% CI: 0.546--0.609) on
NSCLC-Radiomics~\cite{aerts_data_2019}. On
Colorectal-Liver-Metastases~\cite{simpson_preoperative_2023}, CAPTOR ranked second
with an AUC of~0.521~(95\% CI: 0.427--0.604), behind COLIPRI (0.549,
95\% CI: 0.471--0.621) (Figure \ref{fig:bar_plot_tumor}). Although few differences are
statistically significant, the overall ranking pattern was consistent
with the primary classification benchmark.

Computation times for the
Colorectal-Liver-Metastases~\cite{simpson_preoperative_2023} are presented in Appendix \ref{app:results_computation_times}.

\subsection{Ablation of anatomical
preprocessing}\label{ablation-of-anatomical-preprocessing}

CAPTOR significantly outperformed CAPTOR-NoPre, the variant trained on
the same data but without our preprocessing, on the whole test set
(0.849, 95\% CI: 0.828-0.866 vs 0.804, 95\% CI: 0.779-0.824) and on all
test subsets, especially on the CAP subset (0.767, 95\% CI: 0.734-0.793
vs 0.687, 95\% CI: 0.644-0.720) and the oncologic one (0.784, 95\% CI:
0.759-0.805 vs 0.710, 95\% CI: 0.684-0.736) (Figure \ref{fig:bar_plot_classif}).

\subsection{Generalizability}\label{generalizability}

\begin{figure}[t]
    \centering
    \includegraphics[width=\linewidth]{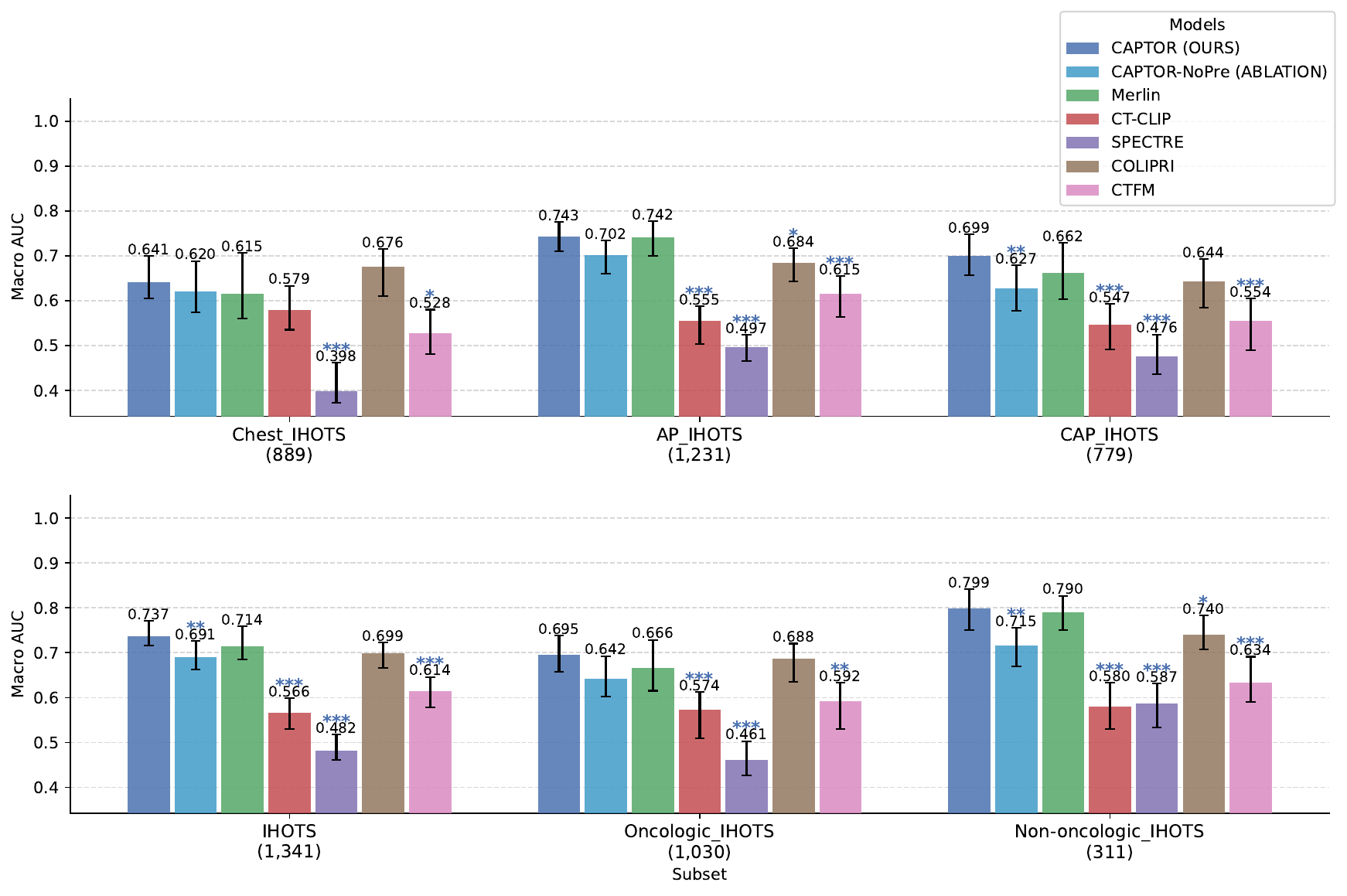}
    \caption{Bar plot showing the classification scores of CAPTOR, CAPTOR-NoPre (i.e. model trained on the same dataset as CAPTOR but without our anatomical crops preprocessing) and five state-of-the-art foundation models, on IHOTS cohort (Institutionally Held-Out Test Subset) stratified by anatomical region visible and indication. One XGBoost was trained for each medical condition and model using the image embedding extracted with the model’s image encoder. For each subset and model, the macro-AUC is the unweighted mean of the AUC scores for each medical condition relevant to the subset among the 26 conditions we studied. The confidence intervals and the pairwise comparison were obtained with a patient-level bootstrap and a patient-level permutation test respectively, both of size 2,000.  CAP=Chest-Abdomen-Pelvis; AUC = Area Under the Receiver Operating Characteristic Curve.  *** = p < 0.001, ** = p < 0.01, * = p < 0.05.}
    \label{fig:IHOTS}
\end{figure}

On the IHOTS test subset, comprising 1,341 examinations acquired at
institutions not represented in CAPTOR\textquotesingle s training data,
CAPTOR achieved a macro-AUC of 0.737 (95\% CI: 0.715-0.771) compared
with 0.714 (95\% CI: 0.685-0.759, Merlin~\cite{blankemeier_merlin_2026}) for the
second best, although the difference is not statistically significant.
Notably, CAPTOR significantly outperformed CAPTOR-NoPre on the whole
IHOTS cohort (0.691, 95\% CI: 0.662-0.726 for CAPTOR-NoPre), supporting
the contribution of the preprocessing strategy to classification
performance on this cohort. The complete stratified results are reported
in Figure \ref{fig:IHOTS}.

Regarding generalizability to external non-oncologic chest dataset,
CAPTOR was significantly outperformed by SPECTRE~\cite{claessens_scaling_2025} and
COLIPRI~\cite{wald_comprehensive_2026}: CAPTOR reached an AUC of 0.796 (95\% CI:
0.785-0.803) on CT-RATE~\cite{hamamci_generalist_2026} vs. 0.847 (95\% CI:
0.836-0.854) for COLIPRI and an AUC of 0.689 (95\% CI: 0.669-0.706) on
RAD-ChestCT~\cite{draelos_rad-chestct_2020} vs. 0.737 (95\% CI: 0.719-0.752) for
COLIPRI (Appendix \ref{app:results_external_datasets}).

\subsection{Effect of training data scale and
composition}\label{effect-of-training-data-scale-and-composition}

\textbf{Medical condition classification}

CAPTOR-100K, trained on the full pretraining dataset (internal and
public image-report datasets), demonstrated comparable performance to
CAPTOR on all subsets of the classification task, the biggest
differences being for the CAP subset (0.744, 95\% CI: 0.703-0.779 vs.
0.767, 95\% CI: 0.734-0.793 for CAPTOR), and the oncologic subset
(0.771, 95\% CI: 0.743-0.796 vs. 0.784, 95\% CI: 0.759-0.805 for
CAPTOR), although they were not significant.

\vspace{5mm}
\textbf{Retrieval}

For retrieval, CAPTOR-100K outperformed all baseline models on the whole
test set, Chest, AP and non-oncologic subsets, reaching a Recall@10 for
the whole test set of 310 (95\% CI:302--319) and MAP@10 of 1.50 (95\%
CI: 1.47-1.54), but was outperformed by CAPTOR on the CAP and oncologic
subsets.

\vspace{5mm}
\textbf{Survival prediction
benchmark}\label{survival-prediction-benchmark-2}

CAPTOR outperformed CAPTOR-100K on the two abdomen datasets of the four
oncologic test cohorts while CAPTOR-100K demonstrated superior
performance on the two chest CT datasets (CAPTOR-100K AUC: 0.616, 95\%
CI: 0.581-0.642 on NSCLC-Radiomics~\cite{aerts_data_2019} and 0.652, 95\%
CI: 0.583-0.689 on NSCLC-Radiogenomics~\cite{bakr_data_2017}), though none
of the differences between the models were statistically significant.

\vspace{5mm}
\textbf{Generalizability}

On the IHOTS test subset, CAPTOR-100K significantly outperformed all
state-of-the-art models. Compared to CAPTOR, it achieved numerically
higher AUC on all subsets except for the non-oncologic one with no
statistical significance, reaching a macro-AUC of 0.752 (95\% CI:
0.726-0.786) on the full cohort compared with 0.737 (95\% CI:
0.715-0.771) for CAPTOR. On the oncologic subset of the IHOTS cohort,
CAPTOR-100K achieved a macro-AUC of 0.717 (95\% CI: 0.677-0.764)
compared with 0.695 (95\% CI: 0.657-0.738) for CAPTOR. On the CAP
subset, the corresponding values were 0.720 (95\% CI: 0.668-0.769) and
0.699 (95\% CI: 0.657-0.749), respectively for CAPTOR-100K and CAPTOR.

Table \ref{Comparison_30_100} summarizes the performance differences between CAPTOR and
CAPTOR-100K across all downstream tasks. Detailed figures of CAPTOR-100K
performances compared to baselines are in Appendix \ref{app:results_data_composition}.

\begin{table}[htbp]
\renewcommand{\arraystretch}{1.1}
\centering
\setlength{\tabcolsep}{3pt}
\begingroup
\newcolumntype{C}[1]{>{\centering\arraybackslash}m{#1}}
\small
\resizebox{\linewidth}{!}{%
\begin{tabular}{C{3cm} C{3.2cm}|C{3.4cm} C{3.4cm}}
\toprule
\textbf{\makecell[c]{Datasets \\ \textit{Task}}} & \textbf{Subset} & \textbf{CAPTOR} & \textbf{CAPTOR-100K} \\
\midrule
& Whole & \makecell{\textbf{0.849} {\tiny(0.828--0.866)}} & \makecell{0.846 {\tiny(0.824--0.862)}} \\
& Chest & \makecell{\textbf{0.838} {\tiny(0.789--0.865)}} & \makecell{0.828 {\tiny(0.778--0.860)}} \\
\multirow{6}{=}[22pt]{\makecell[c]{Global test set \\ \textit{Medical condition} \\\textit{classification}}} & AP & \makecell{0.807 {\tiny(0.797--0.816)}} & \makecell{\textbf{0.808} {\tiny(0.798--0.818)}} \\
& CAP & \makecell{\textbf{0.767} {\tiny(0.734--0.793)}} & \makecell{0.744 {\tiny(0.703--0.779)}} \\
& Oncological & \makecell{\textbf{0.784} {\tiny(0.759--0.805)}} & \makecell{0.771 {\tiny(0.743--0.796)}} \\
& Non-oncologic & \makecell{\textbf{0.859} {\tiny(0.833--0.879)}} & \makecell{0.856 {\tiny(0.831--0.874)}} \\
\midrule
\midrule
& NSCLC-Radiomics & \makecell{0.587 {\tiny(0.546--0.609)}} & \makecell{\textbf{0.616} {\tiny(0.581--0.642)}} \\
\multirow{4}{=}[9pt]{\makecell[c]{Oncological test sets \\ \textit{Two-year survival} \\\textit{prediction}}} & NSCLC-Radiogenomics & \makecell{0.640 {\tiny(0.567--0.676)}} & \makecell{\textbf{0.652} {\tiny(0.583--0.689)}} \\
& C4KC-KiTS & \makecell{\textbf{0.637} {\tiny(0.523--0.697)}} & \makecell{0.556 {\tiny(0.447--0.633)}} \\
& Colorectal-Liver Metastases & \makecell{\textbf{0.521} {\tiny(0.427--0.604)}} & \makecell{0.482 {\tiny(0.403--0.576)}} \\
\midrule
\midrule
& Whole & \makecell{139 {\tiny(125--153)}} & \makecell{\textbf{310} {\tiny(302--319)}} \\
& Chest & \makecell{64.3 {\tiny(51.9--84.0)}} & \makecell{\textbf{95.2} {\tiny(86.2--101)}} \\
\multirow{6}{=}[22pt]{\makecell[c]{Global test set \\ \textit{Text-Image retrieval}}} & AP & \makecell{122 {\tiny(116--129)}} & \makecell{\textbf{254} {\tiny(245--262)}} \\
& CAP & \makecell{\textbf{47.8} {\tiny(44.2--51.6)}} & \makecell{38.6 {\tiny(35.3--41.9)}} \\
& Oncological & \makecell{\textbf{83.4} {\tiny(78.4--88.6)}} & \makecell{71.7 {\tiny(67.3--76.6)}} \\
& Non-oncologic & \makecell{90.3 {\tiny(80.5--101)}} & \makecell{\textbf{265} {\tiny(255--276)}} \\
\midrule
\midrule
& Whole & \makecell{1.46 {\tiny(1.44--1.50)}} & \makecell{\textbf{1.50} {\tiny(1.47--1.54)}} \\
& Chest & \makecell{1.23 {\tiny(1.20--1.27)}} & \makecell{1.24 {\tiny(1.21--1.28)}} \\
\multirow{6}{=}[22pt]{\makecell[c]{Global test set \\ \textit{Image-Image retrieval}}} & AP & \makecell{1.43 {\tiny(1.40--1.46)}} & \makecell{\textbf{1.46} {\tiny(1.44--1.49)}} \\
& CAP & \makecell{\textbf{1.30} {\tiny(1.25--1.35)}} & \makecell{1.28 {\tiny(1.23--1.33)}} \\
& Oncological & \makecell{\textbf{1.41} {\tiny(1.37--1.46)}} & \makecell{1.39 {\tiny(1.35--1.44)}} \\
& Non-oncologic & \makecell{1.41 {\tiny(1.37--1.45)}} & \makecell{\textbf{1.47} {\tiny(1.42--1.51)}} \\
\bottomrule
\end{tabular}
}%
\endgroup
\caption{Effect of training-data volume on CAPTOR performance. Comparison of CAPTOR and CAPTOR-100K performances across all downstream tasks with 95\%confidence intervals in parentheses. Retrieval results are shown for K=10, with K the number of retrieved images considered.}
\label{Comparison_30_100}
\end{table}

CAPTOR-100K statistically significantly outperformed CAPTOR on external
non-oncologic chest CT datasets, reaching an AUC of 0.820 (95\% CI:
0.813-0.827) on CT-RATE~\cite{hamamci_generalist_2026} and 0.719 (95\% CI:
0.698-0.735) on RAD-ChestCT~\cite{draelos_rad-chestct_2020}, while still
outperformed by SPECTRE~\cite{claessens_scaling_2025} and
COLIPRI~\cite{wald_comprehensive_2026}, however not statistically significantly
(Appendix \ref{app:results_external_datasets}).

\vspace{5mm}
\textbf{Classifier training data volume}

We further evaluated the effect of training data volume on classifier
performance by varying the fraction of the training set used during
XGBoost training (Appendix \ref{app:results_data_composition}, Figure \ref{fig:ablation}). Increasing the amount of training data generally leads to improved
performance, although the magnitude and consistency of these gains vary
across datasets. CAPTOR consistently achieves the highest mean AUC
across all training fractions compared to baselines on the whole test
set, the CAP and the oncologic subsets, demonstrating strong robustness
in low-data settings.

\section{Discussion}\label{discussion}

We hypothesized that region-centered anatomical preprocessing would
improve whole-examination representation learning of CAP CT performed in
oncologic settings, beyond the contribution of the training data alone,
with the clinical goal of detecting conditions relevant in an oncologic
setting. Our four experiments support this hypothesis, show that it
holds at unseen institutions, and indicate that broader pretraining data
may add robustness outside the training distribution.

\textbf{Comparison with CT foundation models:}~CAPTOR achieved the
highest classification and retrieval performance on the CAP and
oncologic subsets, outperforming all five baselines, including
generalist models pretrained on substantially larger datasets. CAPTOR
ranked first on three of the four oncologic survival benchmarks,
although overall discrimination was modest (AUCs ranging from 0.52 to
0.64) and few pairwise differences were statistically significant. These
exploratory results suggest that the learned representations capture
some prognostic information, but do not establish a consistent advantage
over competing models for survival prediction.

\textbf{Ablation of preprocessing:}~CAPTOR-NoPre, trained on the same
data and with the same architecture but without anatomical crops, scored
lower than CAPTOR on the whole test set (0.804 vs 0.849) and on the CAP
(0.687 vs 0.767) and oncologic (0.710 vs 0.784) subsets. Without
preprocessing, a model trained on 25,512 predominantly oncologic
examinations performed no better than Merlin~\cite{blankemeier_merlin_2026}, which
was trained on largely non-oncologic abdominal-pelvic scans. These
results demonstrate that the proposed anatomical preprocessing strategy
contributes substantially to classification performance, independently
of the training dataset.

\textbf{Institutional generalization:}~On IHOTS (1,341 examinations from
56 institutions absent from training), CAPTOR remained numerically ahead
of all five baselines and of CAPTOR-NoPre (0.737 vs 0.691), although its
margin over Merlin~\cite{blankemeier_merlin_2026} and COLIPRI~\cite{wald_comprehensive_2026}
(0.714 and 0.699 respectively) was not significant. Nevertheless, the
benefit of preprocessing held across institutions.

\textbf{Training data scale and composition:}~CAPTOR-100K matched CAPTOR
on the CAP and oncologic subsets, with no significant differences. It
was numerically higher on IHOTS and significantly higher on
CT-RATE~\cite{hamamci_generalist_2026} and RAD-ChestCT~\cite{draelos_rad-chestct_2020}.
CAPTOR-100K significantly outperformed all state-of-the-art models on
IHOTS, whereas the differences between Merlin~\cite{blankemeier_merlin_2026} and
COLIPRI~\cite{wald_comprehensive_2026} and CAPTOR were not significant. These
findings suggest that broader pretraining may provide advantages when
evaluating models across heterogeneous institutions and imaging
populations, although the absence of significant differences between
CAPTOR and CAPTOR-100K on IHOTS prevents establishing a clear advantage.
Furthermore, because the two models differ in both training dataset size
and composition, the independent contributions of these factors remain
undetermined.

This study has several limitations. First, this is a retrospective
study, with no prospective validation. Second, reference labels were
automatically extracted from radiology reports using a large language
model rather than manual expert annotation. The reliance on
report-derived labels rather than large-scale expert annotations
reflects a broader challenge in medical imaging foundation model
development, where obtaining high-quality expert annotations at scale
remains difficult. Although this approach enables training and
evaluation on substantially larger datasets, and performance was
evaluated on the public dataset CT-RATE~\cite{hamamci_generalist_2026}, omissions
in the original reports and errors in automated label extraction may
introduce label noise. Finally, CAPTOR is a case-level foundation model:
it aggregates information across large anatomical regions into a single
study-level embedding, enabling efficient inference and strong
examination-level performance, but is not designed for lesion-level
localization or characterization. Architectures such as
SPECTRE~\cite{claessens_scaling_2025}, which preserve fine-grained spatial
information, remain better suited to lesion-level tasks, and our results
should be interpreted within the context of global examination
understanding rather than lesion analysis.

In conclusion, region-centered anatomical preprocessing improved
whole-examination representation learning for CAP CT, including at
institutions unseen during training. This approach enabled a
region-specific vision-language architecture to capture clinically
relevant information across oncologic and non-oncologic findings, while
achieving performance comparable to a substantially larger-scale model.
Broader pretraining may offer additional benefits under distribution
shift, although the respective contributions of training data scale and
composition remain to be established. Code and model weights are
publicly available.

\section{Acknowledgments}\label{Acknowledgments}
Research study funded by Guerbet and BPI France within the scope of the France 2030 project.
This work was performed using HPC resources from GENCI–IDRIS (2026-103370).

\bibliographystyle{unsrtnat}
\bibliography{references}

\clearpage
\newpage
\appendix
\section{Supplementary methods}
\subsection{Internal dataset}\label{app:methods_internal_dataset}
For each study, series were selected via a priority rule favoring full-coverage and contrast-enhanced acquisitions. The series selection priority is the following: CAP > Chest + AP > CA + Pelvis > Chest + Abdomen + Pelvis > Unknown > AP > CA > Chest or Abdomen or Pelvis. If several series correspond to a category (several CAP series for example), we select the series based on the phase and kernel type : Venous + Smooth > Unenhanced + Smooth > Delayed + Smooth > Arterial + Smooth > Unknown + Sharp > Unknown + Smooth > Venous + Sharp > First available series.

\subsection{External dataset series selection criteria} \label{app:methods_image_text_dataset}
INSPECT was excluded from the testing set because the reports only include the Impressions section which is not detailed enough to extract reliable labels, and BIMCVR was excluded from the test set due to its limited size and lack of descriptive information. 

\subsection{More information on oncologic datasets} \label{app:methods_oncologic_dataset}
The four public oncologic datasets we used among those in TumorImagingBench are the following :

-	NSCLC-Radiomics~\cite{aerts_data_2019} : CT scans from 421 non-small cell lung cancer (NSCLC) patients, who received radiation therapy or chemo-radiation therapy. Pretreatment CT scans, expert segmentation of Gross Tumor Volume (GTV) and clinical outcome data are available for each patient. The median survival time of this study is 17.9 months.

-	NSCLC-Radiogenomics~\cite{bakr_data_2017} : Cohort of 211 NSCLC patients with preoperative CT and PET/CT scans. We used images from the 133 patients with GTV annotated like TumorImagingBench. The median survival time of these patients is 44.5 months.

-	C4KC-KiTS~\cite{heller_c4kc_2019} : CT scans and tumor and kidney segmentations from renal tumor patients with partial or radical nephrectomy. Of the 210 patients in the dataset, 134 remain after filtering for survival data at 2 years. The median survival time of these patients is 15.5 months.

-	Colorectal-Liver-Metastases~\cite{simpson_preoperative_2023} : Preoperative CT scans and liver/tumor/vessels segmentation from 194 patients who underwent colorectal liver metastases (CRLM) resection. The median survival time of this study is 44.5 months.

We did not use the two other datasets originally present in TumorImagingBench because they are lesion-level, which is not our aim with CAPTOR.

\subsection{More information on preprocessing}\label{app:methods_preprocessing}
CT volumes were reformatted to RAS orientation, resampled to isotropic in-plane resolution (1.5 mm) and 3-mm slice spacing, intensity-normalized, padded, and cropped to a fixed input size of 224x224x160 voxels.
The full list of regions we used is the same as Merlin: chest, liver and biliary tree, gallbladder, spleen, pancreas, adrenal glands, kidneys and ureters, gastrointestinal tract, peritoneal cavity, pelvic organs, vasculature and lymph nodes, musculoskeletal

Regarding report preprocessing, reports were automatically checked to make sure they did not contain unreadable characters. 94\% of reports were acquired already in English (originally in English or already translated); a smaller subset (1,805 studies, 6\% of the dataset) remained in Spanish and was used as-is for training, without separate handling. In the test set, Spanish reports represented a smaller share (2\%) and were all excluded from all analyses.

\subsection{Training hyperparameters} \label{app:methods_training}
We used an Adam optimizer with 1.25e-6 learning rate, $\beta = (0.9, 0.99)$ and used gradient checkpointing for both the visual and text encoders. We used a batch size of 18 on a single H100 GPU.

The exact same training hyperparameters and validation method are used to train CAPTOR, CAPTOR-100K and CAPTOR-NoPre

\subsection{Baseline Models}\label{app:methods_baselines}
We compared our model with five state-of-the-art foundation models. Two of them were trained on region-specific CT datasets:

-	MERLIN~\cite{blankemeier_merlin_2026}:  3D contrastive vision-language model based on an inflated ResNet-152 image encoder and a Clinical Longformer text encoder. It was trained on approximately 25,000 abdomen-pelvis CT-report pairs from the STARR repository (Merlin dataset), consisting predominantly of non-oncologic examinations.

-	CT-CLIP~\cite{hamamci_generalist_2026}: a 3D contrastive vision-language model using transformer vision and text encoders (adapted from CT-ViT for the vision encoder and CXR-Bert for the text encoder), trained on the CT-RATE dataset comprising approximately 25,000 non-oncologic chest CT-report pairs. 

The others are trained on broader datasets: 

-	SPECTRE~\cite{claessens_scaling_2025}: a 3D Vision Transformer (ViT) pretrained using a combination of self-supervised and contrastive vision-language objectives. It was trained on 229,619 CT series from multiple public datasets spanning chest, abdomen, pelvis, and whole-body imaging (including CT-RATE, MERLIN, INSPECT, NLST, AMOS, AbdomenAtlas, PANORAMA, and AbdomenCT-1K). Although anatomically diverse, the training corpus was largely non-oncologic, with pancreatic cancer data (PANORAMA) representing approximately 1\% of the total dataset.

-	COLIPRI~\cite{wald_comprehensive_2026}: a transformer-based multimodal foundation model combining contrastive vision-language and vision-only pretraining. Foundation model training used CT-RATE~\cite{hamamci_generalist_2026} and NLST and downstream tasks used the same public CT-report datasets included in CAPTOR-100K, complemented by large image-only MRI datasets (brain, knee/thigh, and whole-body MRI), making it the broadest multimodal medical foundation model evaluated (in terms of the intended use-case of the model).

-	CT-FM~\cite{pai_vision_2025}: a vision-only 3D foundation model trained using contrastive self-supervised learning on more than 148,000 CT examinations acquired across multiple cancer types and associated comorbidities. Unlike the other vision-language models, CT-FM does not use radiology reports during pretraining and learns image representations exclusively from CT volumes.

Table \ref{Baseline_models} summarizes their characteristics. 
\begin{table}[htbp]
\centering
\renewcommand{\arraystretch}{1.4}
\begin{adjustbox}{max width=0.45\textwidth}
\begin{tabular}{lccccc}

\toprule
\textbf{Model}
  & \textbf{VLM}
  & \textbf{Body Parts}
  & \textbf{Training Dataset(s)}
  & \textbf{Dataset size}   \\
\midrule

\makecell[c]{CAPTOR \\ (OURS)}
  & \checkmark
  & CAP
  & Internal 
  & 30K \\
\midrule


Merlin
  & \checkmark
  & Abdomen \& Pelvis
  & Merlin
  & 25K\\
\midrule

CT-CLIP
  & \checkmark
  & Chest
  & CT-RATE 
  & \makecell[c]{50K volumes \\ 25K scans}\\
\midrule

SPECTRE
  & \checkmark
  & CAP
  & \makecell[c]{Multiple open CT datasets incl.\\ CT-RATE, INSPECT, Merlin}
  & 230K\\
\midrule

COLIPRI
  & \checkmark
  & Chest
  & CT-RATE, NLST 
  & 100K\\

\midrule
CT-FM
  & \xmark
  & CAP + others 
  & \makecell[c]{Imaging Data Commons \\ (not used in CAPTOR training)}
  & 148K \\

\bottomrule
\end{tabular}
\end{adjustbox}
\caption{Characteristics description of CAPTOR and the five baseline models used in analyses. VLM = Vision-Language Model. Merlin~\cite{blankemeier_merlin_2026}, CT-RATE~\cite{hamamci_generalist_2026} and INSPECT~\cite{huang_inspect_2023} datasets were used in CAPTOR-100K and classifier training for disease detection, Merlin~\cite{blankemeier_merlin_2026} and CT-RATE’s~\cite{hamamci_generalist_2026} test sets were included in our test set.}
\label{Baseline_models}
\end{table}

To compare the embedding quality of each model, we used the official documentation and code to preprocess the data and extract embeddings. For CAPTOR and the Merlin model we used the embeddings from the second to last layer, as it is the official image embedding method.

\subsection{Label extraction - Prompt and conditions}\label{app:methods_label_extraction}
We used zero-shot prompting and intentionally avoided few-shot prompting because selecting representative demonstration examples introduces an additional source of variability and may reduce reproducibility across datasets.
The prompt we used for the label extraction with Mistral is the following: 

\textbf{System}: "You are a radiologist reviewing radiology reports and classifying whether the patient has a condition or not."

\textbf{Prompt}: 
“Here is a radiology report from what radiologists can see in CT scans: \{report\_content\} and here is a list of medical conditions that can potentially be stated in these reports: \{list\_conditions\} 
You are to determine if:

- Each condition is stated or explained as detected in the report. In this case you will put the string 'True' in the field 'detected' of the right condition, the string 'False' otherwise.

- Each condition is stated or explained as not detected in the report. If the report states that there is a history of the condition in the patient, that counts as not detected condition. If the condition is stated as not detected in the report you will put the string 'True' in the field 'not\_detected' in the field of the right condition, the string 'False' otherwise

- Each condition is not stated or explained at all in the report. In this case, you will put the string 'True' in the field 'not\_mentioned' of the field of the right condition, the string 'False' otherwise.
You will give the results as a JSON with the following schema \{json\_schema\}, with strings 'True' or 'False' in the fields 'detected', 'not\_detected', and 'not\_mentioned' for each condition.

If the condition is stated as detected but the report also states that there is a history of this condition in the patient, please fill 'False' in the 'not\_detected' field because a condition cannot be present and not present at the same time.
There can only be one 'True' for each condition.
Please do not add any condition that is not in the list I gave you.”

The full list of conditions is the following : Arterial dissection, Hemoperitoneum, Mesenteric ischemia, Active bleeding, Infarction of solid organs, Pseudo aneurysm, Visceral perforation, Pneumoperitoneum (free air), Peritonitis, Pyelocaliceal dilation, Arterial/venous occlusion, Fracture, Pancreatitis, Aortic aneurysm (unruptured), Intestinal obstruction, Liver fibrosis / Cirrhosis, Peritoneal carcinomatosis (omental caking), Pneumothorax, Pulmonary embolism, Pericardial effusion, Epiduritis, Pleural effusion, Peritoneal effusion, Intrahepatic bile duct dilatation, Common bile duct dilatation, Coronary artery disease \/ calcification

\subsection{XGBoost parameters and details on classification method}\label{app:methods_classif_details}
XGBoost was chosen as a strong non-linear classifier while using identical hyperparameters across all conditions and foundation models, ensuring that performance differences primarily reflected embedding quality rather than downstream model optimization.
To train the XGBoost classifiers the binary logistic objective function (objective = binary:logistic) was used for classification. To account for the unequal class distribution, the positive class was weighted using the scale\_pos\_weight parameter. A moderate learning rate (learning\_rate = 0.1) was selected, and model complexity was constrained by limiting tree depth (max\_depth = 3) and by applying both L1 (reg\_alpha = 0.5) and L2 (reg\_lambda = 5) regularization. Additional variance reduction was introduced through row subsampling (subsample = 0.9) and feature subsampling (colsample\_bytree = 0.9). The number of boosting iterations was fixed at 100 trees (n\_estimators = 100). Overall, these settings were chosen to favor model generalization and robustness while limiting overfitting, particularly in datasets containing a limited number of positive samples.

For the baseline models we also used all available series, using each model’s method to get the embeddings from each series and using the embeddings from the right series for each condition (e.g. embeddings from the Chest series for chest conditions and from the AP series for AP conditions).

\subsection{Further explanation on generalization evaluation}\label{app:methods_generalization}
We tested all models on broadly used chest CT datasets: CT-RATE~\cite{hamamci_generalist_2026} and RAD-ChestCT~\cite{draelos_rad-chestct_2020}. 

The original RAD-ChestCT dataset contains 35,747 non-contrast chest CT scans from 19,661 adult patients of the Duke University Health System between January 2012 and April 2017. The publicly available version contains 3,630 chest CT scans, approximately 10\% of the dataset. Each CT scan is annotated with a matrix of 84 abnormality labels x 52 location labels. For our use case we only used the 84 abnormality labels: if the abnormality is present in at least one location, then we set it to present, absent otherwise. 

For both datasets we used the given labels, extracted from the reports with a fine-tuned RadBERT-RoBERTa-4m~\cite{yan_radbert_2022} for CT-RATE and with a rule-based method for RAD-ChestCT, and the given training and testing sets to train and test the classifiers.

\subsection{Further explanation on retrieval metrics}\label{app:methods_retrieval}
Retrieval tasks evaluate embedding-space quality by measuring whether semantically similar images and reports are mapped close together.

For text-image retrieval, the random baseline for Retrieval@K is simply $\frac{K}{N}$, where N is the total number of images, as each image has an equal probability of appearing in the top K retrieved images under random ranking. Consequently, both the random baseline and model scores increase with K, since the probability of retrieving the correct image within the top K candidates increases as K grows.

For image-image retrieval, following standard practice, the relevancy of an image with the reference image is computed with the Intersection over Union of labels, also known as the Jaccard index, thresholded at 0. The IoU consists in the number of positive labels in common (size of the intersection of labels) over the number of positive labels in total for both images (size of the union of labels). In short, in our computation, an image is considered relevant with another if they have the presence of at least one medical condition (positive label) in common. 
The MAP@K metric is then computed following the standard formulation~\cite{manzhos_average_2026}:

\begin{equation}
  AP@K = \frac{1}{min(R,K)}\sum_{i=1}^{K}P@i\cdot\mathbf{1}[rel(i)]
\end{equation}

where R is the total number of relevant items among all the images for the query, P@i is the precision at rank i (the proportion of relevant images in the top  i retrieved images), and $\mathbf{1}[rel(i)]$ indicates whether the item retrieved at rank i is relevant. The MAP@K is then the mean of AP@K over all queries. Importantly, only abnormal volumes (i.e., volumes presenting at least one positive label) are included as queries, as normal volumes have no relevant counterpart by construction under the Jaccard threshold of 0. The random image-image retrieval baseline is computed empirically via Monte Carlo simulation (50 trials, 500 queries per trial), where items are retrieved randomly. The random baseline increases from K=1 to K=5 before decreasing for larger K values, which is expected under the standard AP@K formulation. As K increases, the normalization by min(R,K) can initially inflate scores for queries with few relevant items (if the single relevant item is retrieved within K positions, AP@K equals 1 regardless of K) before the dilution of non-relevant items at higher ranks drives the score down. The baseline also varies across dataset subsets, reflecting differences in label density: subsets with more frequent or overlapping pathologies (e.g., Chest) exhibit higher random baselines than sparser subsets (e.g., AP), as a greater proportion of the gallery shares at least one label with any given query. Model performance is consequently expected to follow the same decreasing trend with K, since retrieving relevant items becomes relatively harder as the retrieved set grows. 

\section{Complementary results}
\subsection{Detailed internal dataset characteristics} \label{app:results_internal_demographics}

Table \ref{Demographics_internal_by_country} presents the internal dataset detailed characteristics by country.

\begin{table*}[!htbp]
\centering
\renewcommand{\arraystretch}{1.3}
\begin{adjustbox}{max width=0.8\textwidth}
\begin{tabular}{
  l
  >{\centering\arraybackslash}m{1.8cm}
  >{\centering\arraybackslash}m{2.2cm}
  >{\centering\arraybackslash}m{2cm}
  >{\centering\arraybackslash}m{4cm}
  >{\raggedright\arraybackslash}m{9cm}
}
\toprule
\textbf{Country} & \textbf{Studies} & \textbf{Age Median [IQR]} & \textbf{Sex} & \textbf{Contrast} & \textbf{Description} \\
\midrule
 
France
& 4{,}254
& 68 [56--77]
& \makecell[l]{F = 53\% \\ M = 47\%}
& \makecell[l]{Injected = 67.8\% \\ Non-injected = 13.4\% \\ Unknown = 18.8\%}
& Data collected for lesion detection throughout the CAP (incl. adrenal glands, bladder, kidneys, spleen, lungs, lymph nodes, ovaries, rectum, and colon) \\
 
Japan
& 949
& NA
& \makecell[l]{F = 50\% \\ M = 47\%}
& \makecell[l]{Injected = 66.1\% \\ Non-injected = 21.1\% \\ Unknown = 12.9\%}
& Data collected for incidental findings of pancreatic cancer. \\
 
Spain
& 4{,}645
& 66 [58--75]
& \makecell[l]{F = 47\% \\ M = 53\%}
& \makecell[l]{Injected = 47.5\% \\ Non-injected = 9.4\% \\ Unknown = 43.1\%}
& Data collected for lesion detection throughout the CAP (incl. the adrenal glands, bladder, kidneys, spleen, lungs, lymph nodes, ovaries, rectum, and colon) \\
 
USA
& 19{,}852
& 66 [56--74]
& \makecell[l]{F = 49\% \\ M = 51\%}
& \makecell[l]{Injected = 71.7\% \\ Non-injected = 21.9\% \\ Unknown = 6.3\%}
& Data collected for disease detection on CAP examinations in an oncologic setting and for lesion detection throughout the CAP (incl. bone, adrenal glands, lungs, gastrointestinal tract, lymph nodes, and pancreas) \\
 
Vietnam
& 193
& NA
& \makecell[l]{F = 41\% \\ M = 59\%}
& \makecell[l]{Injected = 80.3\% \\ Non-injected = 13.0\% \\ Unknown = 6.7\%}
& Data collected for incidental findings of pancreatic cancer. \\
 
\bottomrule
\end{tabular}
\end{adjustbox}
\caption{Detailed characteristics of the proprietary dataset by country. IQR = interquartile range, F=female, M=male}
\label{Demographics_internal_by_country}
\end{table*}
\subsection{Mistral performance}\label{app:results_mistral_perfs}
\begin{figure*}[t]
    \centering
    \includegraphics[width=0.7\linewidth]{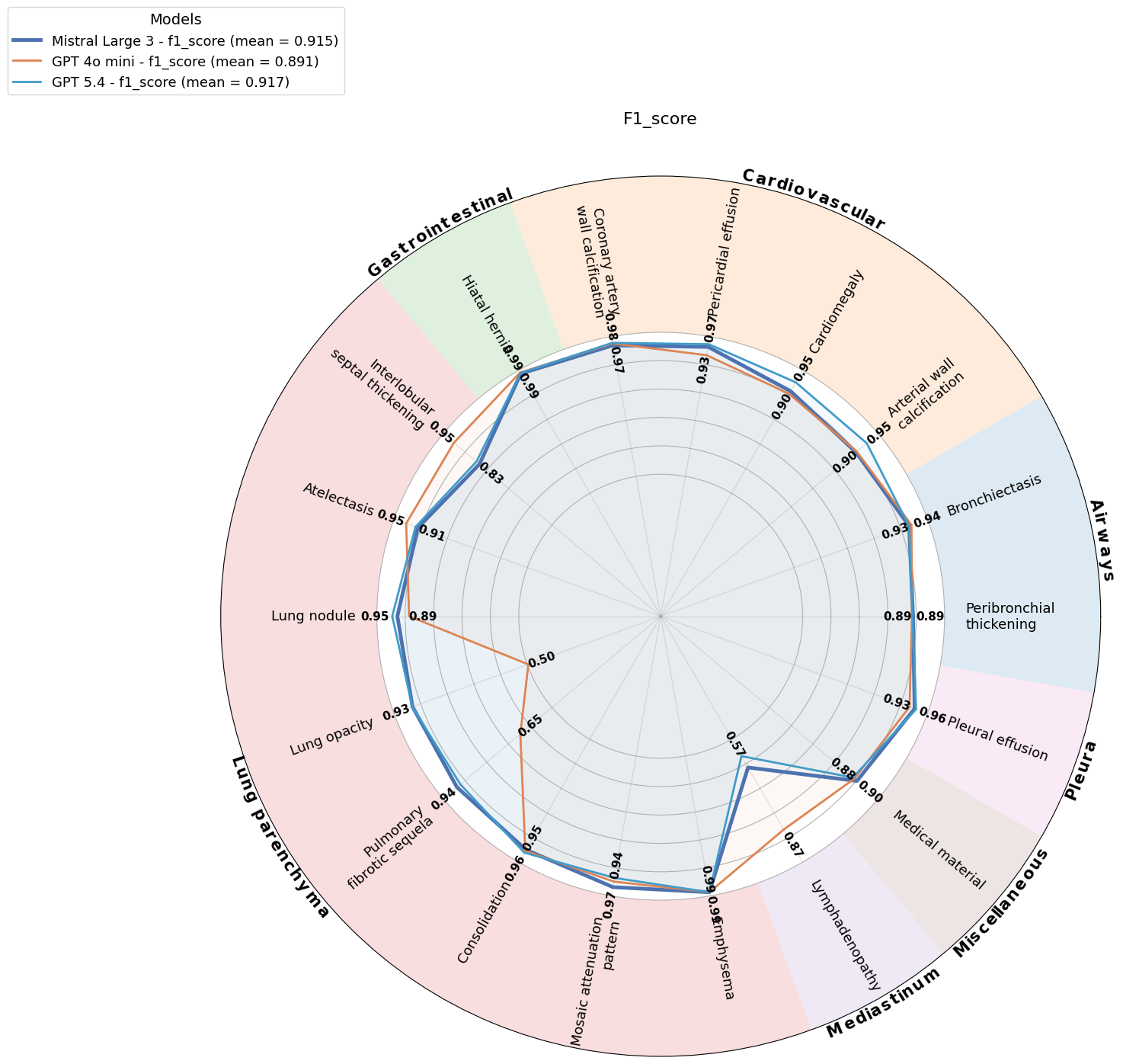}
    \caption{Radar plot comparing the per-condition classification F1-score of Mistral Large 3, GPT-5.4, and GPT-4o mini on the CT-RATE dataset, evaluated against the dataset's original labels (extracted with a fine-tuned RadBERT-RoBERTa-4m on 1,000 reports).}
    \label{fig:radarplot_mistral}
\end{figure*}

Figure \ref{fig:radarplot_mistral} displays the classification performance of Mistral Large 3~\cite{mistral_ai_introducing_nodate}, GPT 5.4~\cite{openai_introducing_2026} and GPT 4o mini~\cite{openai_gpt-4o_nodate} on the CT-RATE~\cite{hamamci_generalist_2026} dataset, using the dataset’s original labels extracted with a fine-tuned RadBERT-RoBERTa-4m~\cite{yan_radbert_2022} on 1000 reports.

\subsection{Label proportions} \label{app:results_label_proportions}
Table \ref{label_proportions} shows the proportions of each medical condition in each subset of our whole testing set (internal dataset combined with public datasets)

\begin{table*}[htbp]

\centering
\renewcommand{\arraystretch}{1.3}
\begin{adjustbox}{max width=\textwidth}
\begin{tabular}{lcccccc}

\toprule
\textbf{Pathology} & \textbf{\makecell[c]{Whole test set \\ N=10,794}} & \textbf{\makecell[c]{Chest \\ N=3,944}} & \textbf{\makecell[c]{AP \\ N=8,731}} & \textbf{\makecell[c]{CAP \\ N=1,881}} & \textbf{\makecell[c]{Oncological \\ N=3,264}} & \textbf{\makecell[c]{Non-\\oncologic \\ N=7,530}} \\
\midrule
Active bleeding & 0.52\% & 0.04\% & 0.73\% & 0.05\% & 0.09\% & 0.68\% \\
Aortic aneurysm (unruptured) & 5.13\% & 9.22\% & 3.13\% & 7.92\% & 6.29\% & 4.72\% \\
Arterial dissection & 0.33\% & 0.19\% & 0.46\% & 0.52\% & 0.39\% & 0.31\% \\
Arterial/venous occlusion & 4.39\% & 1.99\% & 5.99\% & 4.66\% & 4.71\% & 4.28\% \\
Common bile duct dilatation & 4.48\% & - & 6.27\% & 2.48\% & 2.91\% & 5.04\% \\
Coronary artery disease / calcification & 25.05\% & 36.65\% & - & 40.06\% & 35.82\% & 21.19\% \\
Epiduritis & 0.07\% & 0.09\% & - & 0.16\% & 0.12\% & 0.05\% \\
Fracture & 8.54\% & 6.86\% & 10.66\% & 13.30\% & 11.81\% & 7.37\% \\
Hemoperitoneum & 1.07\% & - & 1.51\% & 0.10\% & 0.27\% & 1.35\% \\
Infarction of solid organs & 1.59\% & 0.42\% & 2.23\% & 1.09\% & 0.87\% & 1.85\% \\
Intestinal obstruction & 4.11\% & - & 5.82\% & 0.52\% & 0.99\% & 5.24\% \\
Intrahepatic bile duct dilatation & 6.23\% & - & 8.73\% & 2.85\% & 3.66\% & 7.15\% \\
Liver fibrosis / Cirrhosis & 3.43\% & - & 4.35\% & 2.95\% & 4.92\% & 2.90\% \\
Mesenteric ischemia & 0.60\% & - & 0.86\% & 0.21\% & 0.15\% & 0.76\% \\
Pancreatitis & 1.56\% & - & 2.22\% & 0.78\% & 1.35\% & 1.63\% \\
Pericardial effusion & 5.17\% & 7.13\% & - & 7.04\% & 6.12\% & 4.83\% \\
Peritoneal carcinomatosis (omental caking) & 3.81\% & - & 5.18\% & 8.90\% & 7.97\% & 2.31\% \\
Peritoneal effusion & 16.69\% & - & 22.99\% & 8.18\% & 8.90\% & 19.48\% \\
Peritonitis & 1.94\% & - & 2.76\% & 0.21\% & 0.24\% & 2.55\% \\
Pleural effusion & 14.91\% & 14.69\% & - & 18.22\% & 16.70\% & 14.27\% \\
Pneumoperitoneum (free air) & 2.48\% & - & 3.47\% & 0.26\% & 0.33\% & 3.25\% \\
Pneumothorax & 0.53\% & 0.30\% & - & 0.21\% & 0.15\% & 0.67\% \\
Pseudo aneurysm & 0.23\% & 0.09\% & 0.33\% & 0.26\% & 0.21\% & 0.24\% \\
Pulmonary embolism & 0.51\% & 0.55\% & - & 0.98\% & 0.99\% & 0.34\% \\
Pyelocaliceal dilation & 4.80\% & - & 6.34\% & 4.50\% & 4.38\% & 4.96\% \\
Visceral perforation & 1.54\% & - & 2.18\% & 0.21\% & 0.18\% & 2.03\% \\

\bottomrule
\end{tabular}
\end{adjustbox}
\caption{Medical condition proportions in each subset of our whole testing set (internal dataset combined with public datasets)}
\label{label_proportions}
\end{table*}

\subsection{Statistical significance results by condition}\label{app:results_stats_sig}
Table \ref{nb_path_sig} presents the results of the number of conditions for which the difference between our model CAPTOR and each baseline is statistically significant.
\begin{table*}[htbp]
\centering
\renewcommand{\arraystretch}{1.3}
\begin{adjustbox}{max width=\textwidth}
\begin{tabular}{lcccccc}
\toprule
\textbf{Dataset} & \textbf{\makecell[c]{Number of \\relevant conditions}} & \textbf{Merlin} & \textbf{CT-CLIP} & \textbf{SPECTRE} & \textbf{COLIPRI} & \textbf{CT-FM} \\
\midrule
Whole test set & 26 & +\textbf{8} / -7 & +\textbf{24} / -0 & +\textbf{9} / -4 & +\textbf{20} / -0 & +\textbf{25} / -0 \\
CAP & 26 & +\textbf{11} / -0 & +\textbf{16} / -0 & +\textbf{15} / -0 & +\textbf{13} / -0 & +\textbf{16} / -0 \\
Chest & 13 & +\textbf{9} / -1 & +\textbf{8} / -0 & +\textbf{7} / -0 & +\textbf{3} / -0 & +\textbf{8} / -0 \\
AP & 20 & +\textbf{3} / -6 & +\textbf{19} / -0 & +\textbf{6} / -6 & +\textbf{15} / -0 & +\textbf{19} / -0 \\
Oncological & 26 & +\textbf{13} / -0 & +\textbf{15} / -0 & +\textbf{17} / -0 & +\textbf{15} / -0 & +\textbf{16} / -0 \\
Non-oncologic & 26 & +\textbf{2} / -9 & +\textbf{23} / -0 & +\textbf{4} / -12 & +\textbf{14} / -1 & +\textbf{25} / -0 \\

\bottomrule
\end{tabular}
\end{adjustbox}
\caption{Number of conditions for which the difference between our model CAPTOR and each baseline is statistically significant. '+' shows the number of conditions where CAPTOR is statistically better, '-' shows the number of conditions where the baseline is statistically better.}
\label{nb_path_sig}
\end{table*}

\subsection{Computation times - oncologic datasets}\label{app:results_computation_times}
All the tests for this section were conducted on an Apple MacBook Pro equipped with an Apple M4 Pro chip (12-core CPU: 8 performance cores and 4 efficiency cores) and 24 GB of unified memory. Table \ref{computation_times} presents the computation time for the dataset Colorectal-Liver Metastases, which were consistent with the time on the other datasets. 
\begin{table*}[htbp]
\centering
\renewcommand{\arraystretch}{1.35}
\begin{adjustbox}{max width=\textwidth}
\begin{tabular}{lcc}
\toprule

\textbf{Model}
& \textbf{Total computation time}
& \textbf{Computation time per image}\\
\midrule

CAPTOR (OURS)
& 7min38s
& 2.4s\\

Merlin
& 7min49s
& 2.4s\\

CT-CLIP
& 6min34s
& 2.0s\\

SPECTRE
& 1h34min12s
& 29.1s\\

COLIPRI
& 55min28s
& 17.2s\\

CT-FM
& 1h09min13s
& 21.4s\\
\bottomrule

\end{tabular}
\end{adjustbox}
\caption{Computation times for the dataset Colorectal-Liver Metastases. All computations were conducted on an Apple MacBook Pro equipped with an Apple M4 Pro chip (12-core CPU: 8 performance cores and 4 efficiency cores) and 24 GB of unified memory}
\label{computation_times}
\end{table*}

\subsection{Importance of data composition full results}\label{app:results_data_composition}
This section provides the complete CAPTOR-100K results referenced in the main text, extending the CAPTOR comparisons of Figures \ref{fig:bar_plot_classif}–\ref{fig:bar_plot_tumor} and Table \ref{Retrieval_table} to include the larger training-data	variant.

\begin{figure*}[t]
    \centering
    \includegraphics[width=\linewidth]{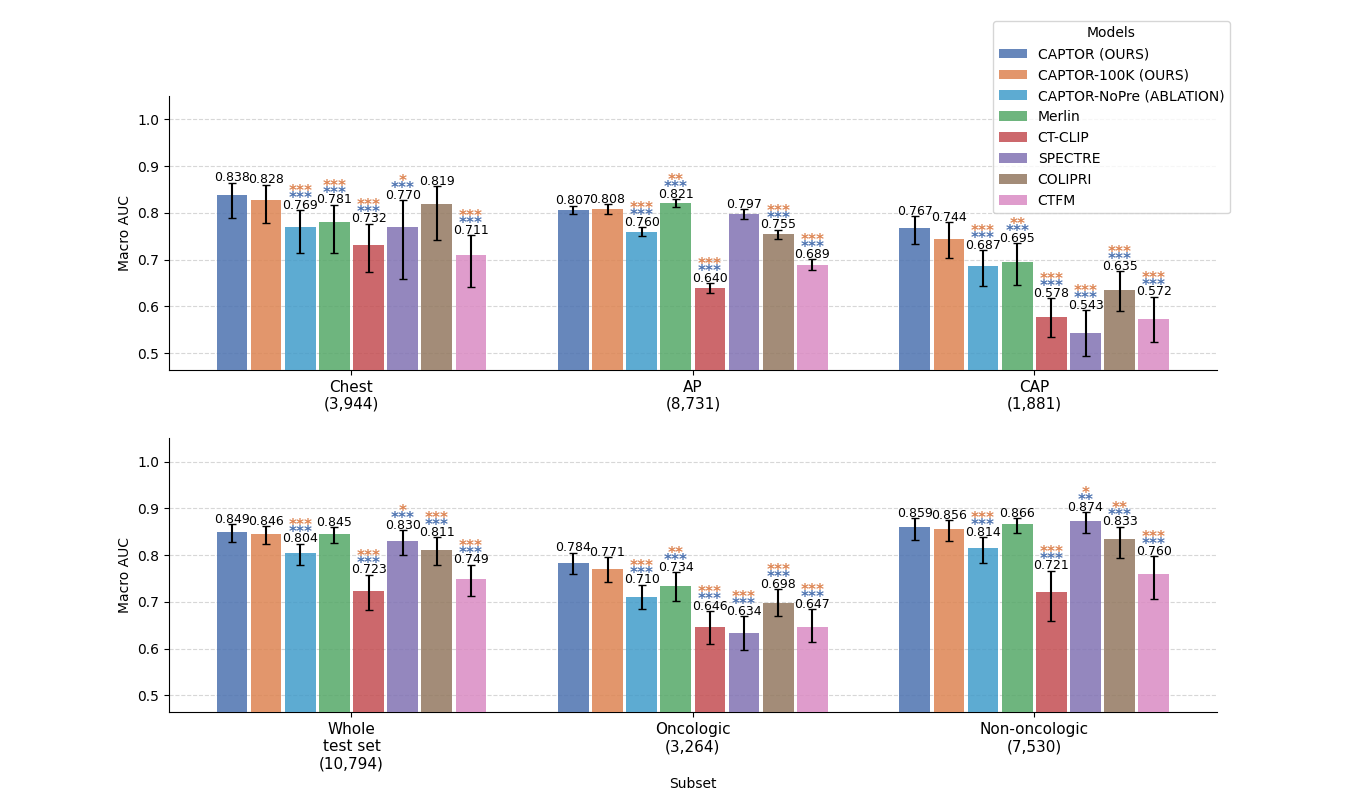}
    \caption{Bar plot showing the classification scores of CAPTOR, CAPTOR-100K, CAPTOR-NoPre (i.e. model trained on the same dataset as CAPTOR but without our anatomical crops preprocessing) and five state-of-the-art foundation models, on our test set stratified by anatomical region visible and indication. One XGBoost was trained for each medical condition and model using the image embedding extracted with the model’s image encoder. For each subset and model, the macro-AUC is the unweighted mean of the AUC scores for each medical condition relevant to the subset among the 26 conditions we studied. The confidence intervals and the pairwise comparison were obtained with a patient-level bootstrap and a patient-level permutation test respectively, both of size 2,000.  *** = p < 0.001, ** = p < 0.01, * = p < 0.05, the stars are the color of the comparison model (orange for CAPTOR-100K and blue for CAPTOR)}
    \label{fig:bar_plot_classif_100K}
\end{figure*}

\begin{figure*}[t]
    \centering
    \begin{subfigure}{0.45\linewidth}
      \centering
      \includegraphics[width=\linewidth]{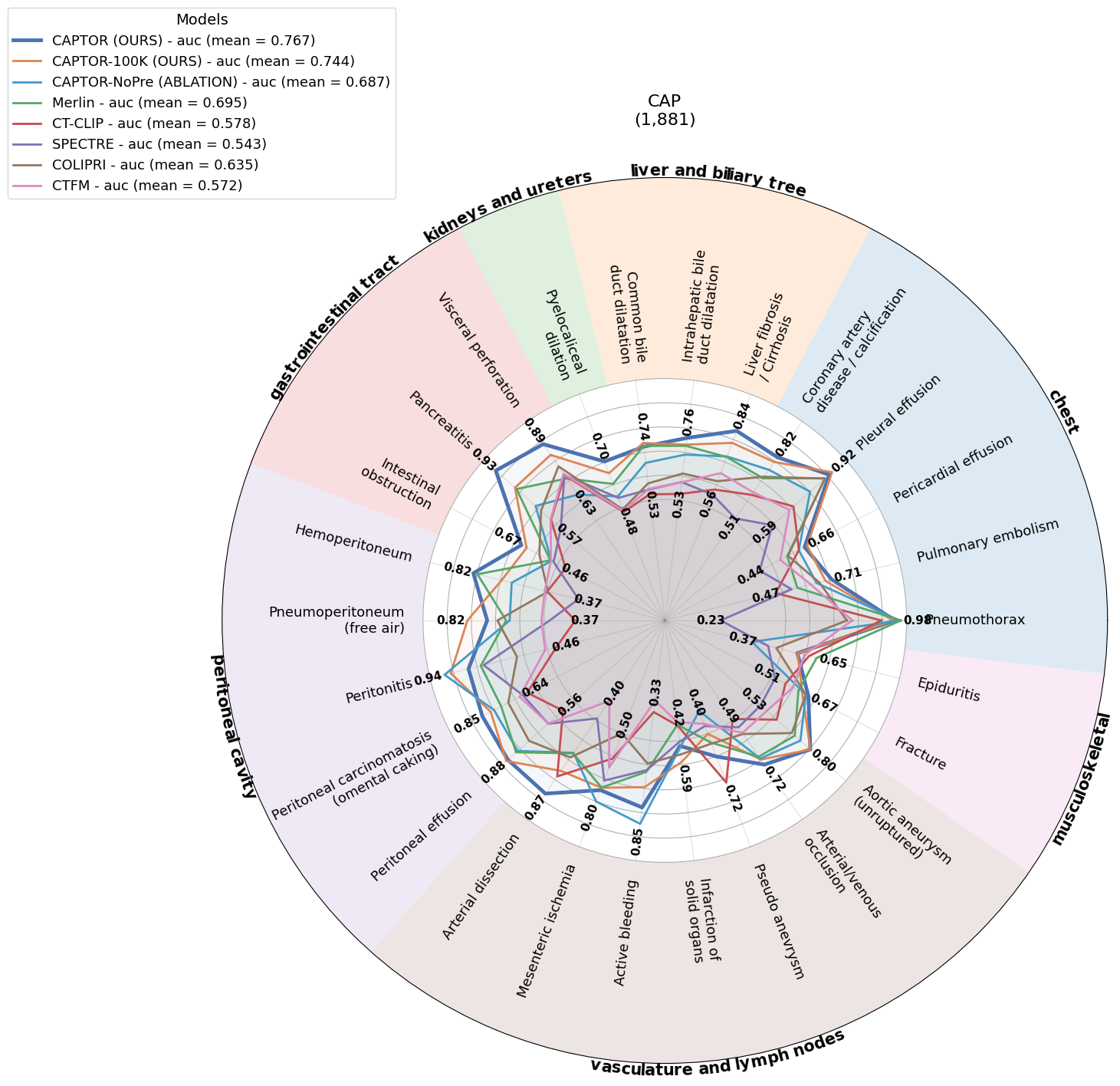}
    \end{subfigure}
    \begin{subfigure}{0.45\linewidth}
      \centering
      \includegraphics[width=\linewidth]{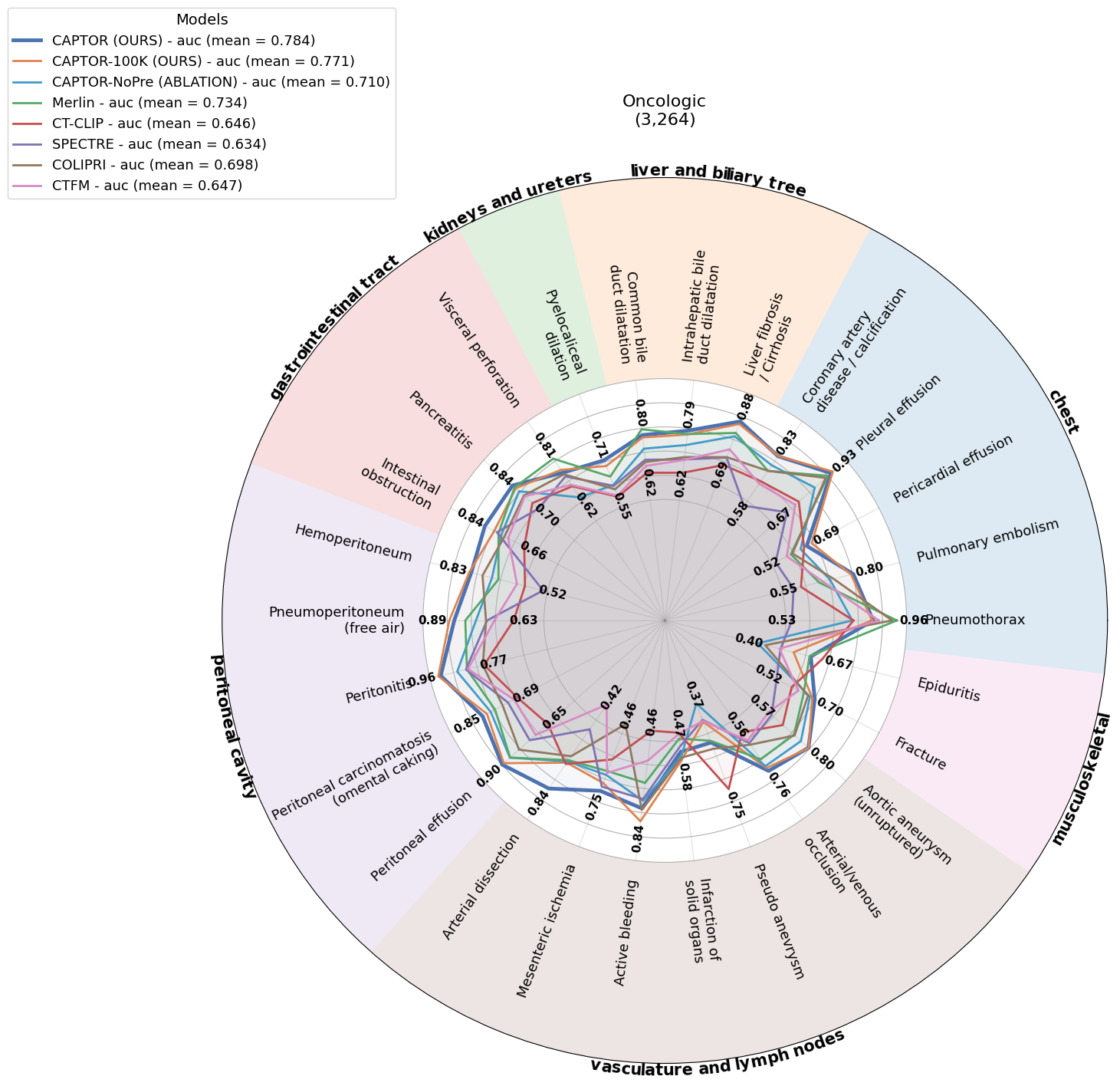}
    \end{subfigure}
    \caption{Per-condition classification performance of CAPTOR, CAPTOR-100K, CAPTOR-NoPre (i.e. model trained on the same dataset as CAPTOR but without our anatomical crops preprocessing), and five state-of-the-art foundation models for the 26 medical conditions, shown separately for the CAP subset (left) and the oncologic subset (right). One XGBoost was trained for each medical condition and model using the appropriate image embedding (i.e. the right regional crop for CAPTOR and the right series for the baselines) extracted with the model’s image encoder}
    \label{fig:radarplot_classif_100K}
\end{figure*}

\begin{figure*}[t]
    \centering
    \includegraphics[width=\linewidth]{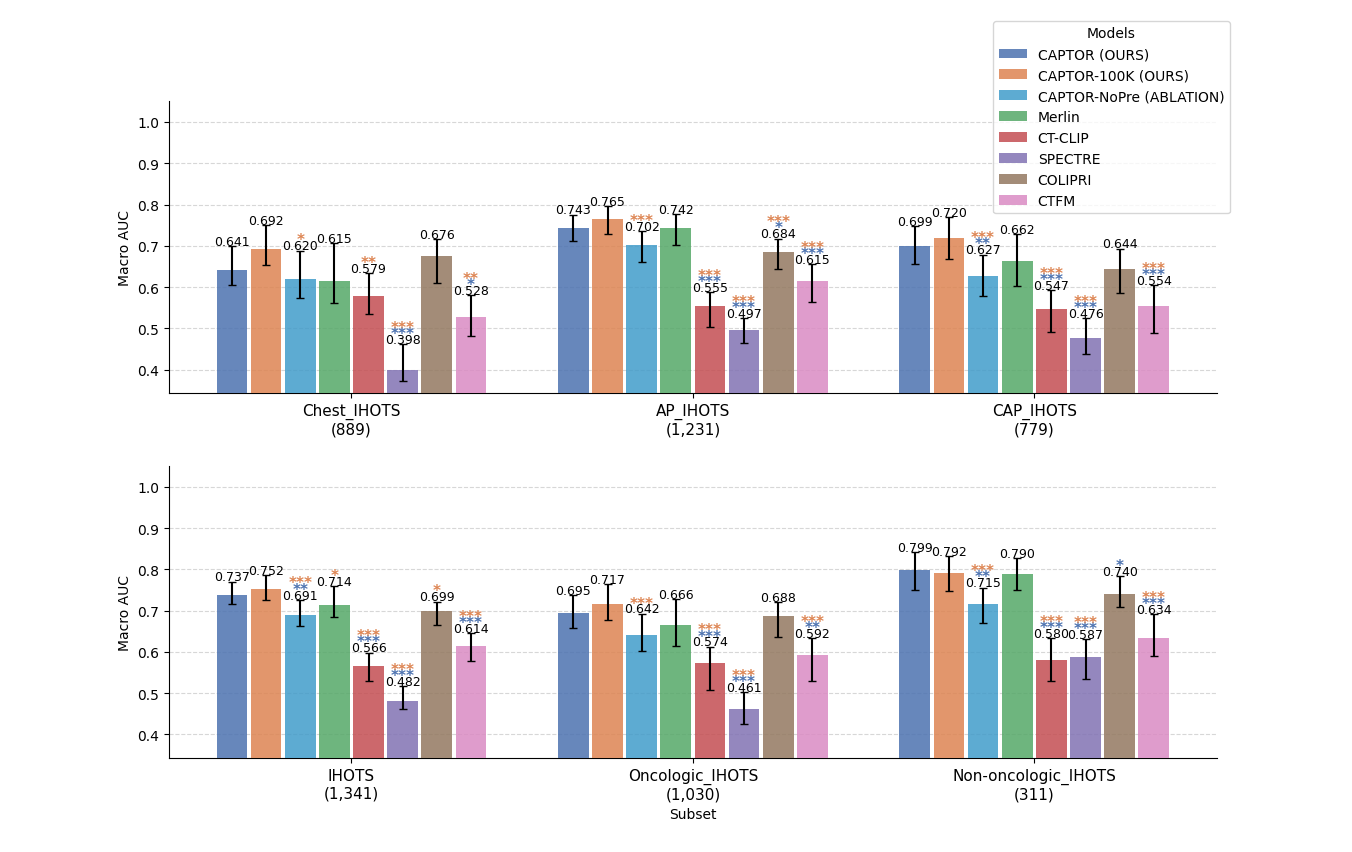}
    \caption{Bar plot showing the classification scores of CAPTOR, CAPTOR-100K, CAPTOR-NoPre (i.e. model trained on the same dataset as CAPTOR but without our anatomical crops preprocessing) and five state-of-the-art foundation models, on IHOTS cohort (Institutionally Held-Out Test Subset) stratified by anatomical region visible and indication. One XGBoost was trained for each medical condition and model using the image embedding extracted with the model’s image encoder. For each subset and model, the macro-AUC is the unweighted mean of the AUC scores for each medical condition relevant to the subset among the 26 conditions we studied. The confidence intervals and the pairwise comparison were obtained with a patient-level bootstrap and a patient-level permutation test respectively, both of size 2,000.  *** = p < 0.001, ** = p < 0.01, * = p < 0.05, the stars are the color of the comparison model (orange for CAPTOR-100K and blue for CAPTOR)}
    \label{fig:bar_plot_classif_100K}
\end{figure*}

\begin{figure*}[t]
    \centering
    \includegraphics[width=\linewidth]{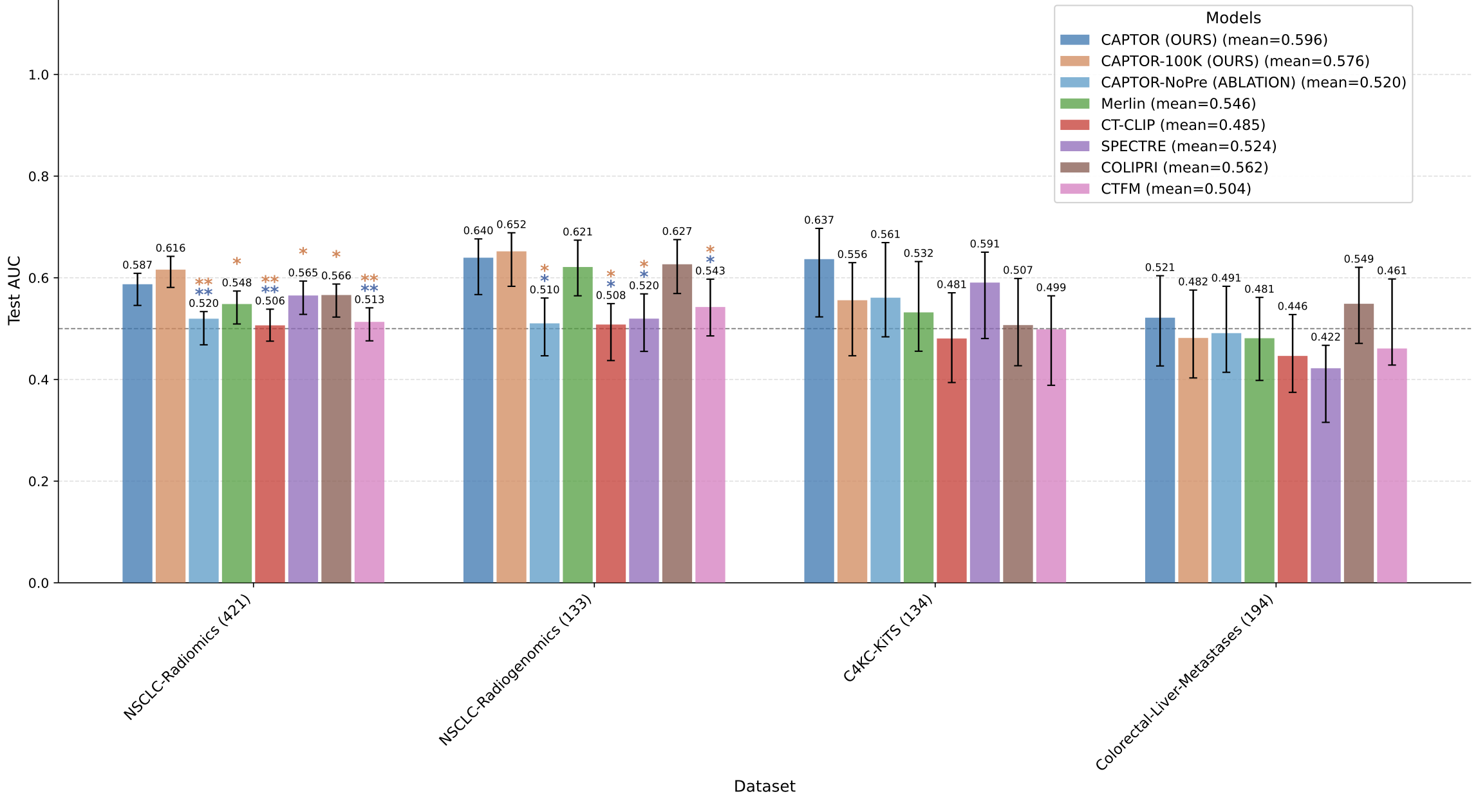}
    \caption{Two-year survival-prediction performance (AUC) of CAPTOR, CAPTOR-100K, CAPTOR-NoPre (i.e. model trained on the same dataset as CAPTOR but without our anatomical crops preprocessing) and five state-of-the-art foundation models on four oncologic TumorImagingBench~\cite{aerts_foundation_2025} cohorts (NSCLC-Radiomics~\cite{aerts_data_2019}, NSCLC-Radiogenomics~\cite{bakr_data_2017}, C4KC-KiTS~\cite{heller_c4kc_2019}, Colorectal-Liver-Metastases~\cite{simpson_preoperative_2023}), using kNN classifiers on frozen image embeddings averaged over 10 random splits.  Statistical significance was assessed using a permutation test (2,000 permutations).}
    \label{fig:bar_plot_tumor_100K}
\end{figure*}

\begin{table*}[htbp]
\renewcommand{\arraystretch}{1.3}
\centering
\begin{adjustbox}{max width=\textwidth}
\begin{tabular}{cc|cccccc|ccccccc}
\toprule
 & & \multicolumn{6}{c|}{Retrieval text-image} & \multicolumn{7}{c}{Retrieval image-image} \\
Dataset & K & \makecell[c]{CAPTOR \\ (OURS)} & \makecell[c]{CAPTOR-100K \\ (OURS)} & \makecell[c]{CAPTOR - NoPre \\ (ABLATION)} & Merlin & CT-CLIP & COLIPRI & \makecell[c]{CAPTOR \\ (OURS)} & \makecell[c]{CAPTOR-100K \\ (OURS)} & \makecell[c]{CAPTOR - NoPre \\ (ABLATION)} & Merlin & CT-CLIP & COLIPRI & CT-FM \\
\midrule
\multirow[c]{5}{*}{\centering \makecell{Whole test set \\ 10,794}} & 1 & \makecell{307 \\{\scriptsize (266--348)}} & \makecell{\textbf{746} \\{\scriptsize (688--797)}} & \makecell{64.0 \\{\scriptsize (47.9--83.0)}} & \makecell{\underline{722} \\{\scriptsize (622--817)}} & \makecell{20.0 \\{\scriptsize (8.88--31.1)}} & \makecell{188 \\{\scriptsize (104--281)}} & \makecell{1.79 \\{\scriptsize (1.73--1.83)}} & \makecell{\textbf{1.88} \\{\scriptsize (1.80--1.91)}} & \makecell{1.72 \\{\scriptsize (1.64--1.74)}} & \makecell{\underline{1.84} \\{\scriptsize (1.77--1.86)}} & \makecell{1.36 \\{\scriptsize (1.28--1.41)}} & \makecell{1.64 \\{\scriptsize (1.55--1.69)}} & \makecell{1.41 \\{\scriptsize (1.34--1.47)}} \\
 & 5 & \makecell{179 \\{\scriptsize (159--199)}} & \makecell{\textbf{437} \\{\scriptsize (421--452)}} & \makecell{52.4 \\{\scriptsize (45.4--60.3)}} & \makecell{\underline{353} \\{\scriptsize (307--394)}} & \makecell{12.2 \\{\scriptsize (6.60--18.7)}} & \makecell{98.4 \\{\scriptsize (57.8--144)}} & \makecell{1.43 \\{\scriptsize (1.38--1.44)}} & \makecell{\textbf{1.47} \\{\scriptsize (1.42--1.48)}} & \makecell{1.38 \\{\scriptsize (1.32--1.38)}} & \makecell{\underline{1.45} \\{\scriptsize (1.40--1.45)}} & \makecell{1.19 \\{\scriptsize (1.12--1.20)}} & \makecell{1.34 \\{\scriptsize (1.27--1.35)}} & \makecell{1.22 \\{\scriptsize (1.15--1.23)}} \\
 & 10 & \makecell{139 \\{\scriptsize (125--153)}} & \makecell{\textbf{310} \\{\scriptsize (302--319)}} & \makecell{45.6 \\{\scriptsize (39.9--51.1)}} & \makecell{\underline{235} \\{\scriptsize (205--261)}} & \makecell{11.3 \\{\scriptsize (6.71--16.6)}} & \makecell{70.1 \\{\scriptsize (42.6--101)}} & \makecell{1.47 \\{\scriptsize (1.45--1.51)}} & \makecell{\textbf{1.51} \\{\scriptsize (1.48--1.55)}} & \makecell{1.41 \\{\scriptsize (1.38--1.44)}} & \makecell{\underline{1.49} \\{\scriptsize (1.46--1.52)}} & \makecell{1.20 \\{\scriptsize (1.17--1.25)}} & \makecell{1.36 \\{\scriptsize (1.32--1.41)}} & \makecell{1.24 \\{\scriptsize (1.21--1.29)}} \\
 & 50 & \makecell{65.8 \\{\scriptsize (60.9--70.4)}} & \makecell{\textbf{119} \\{\scriptsize (117--121)}} & \makecell{29.9 \\{\scriptsize (27.2--32.5)}} & \makecell{\underline{77.1} \\{\scriptsize (67.8--84.8)}} & \makecell{8.02 \\{\scriptsize (4.97--11.8)}} & \makecell{30.7 \\{\scriptsize (21.7--40.5)}} & \makecell{1.60 \\{\scriptsize (1.57--1.65)}} & \makecell{\textbf{1.64} \\{\scriptsize (1.61--1.69)}} & \makecell{1.52 \\{\scriptsize (1.49--1.57)}} & \makecell{\underline{1.61} \\{\scriptsize (1.59--1.66)}} & \makecell{1.24 \\{\scriptsize (1.22--1.31)}} & \makecell{1.43 \\{\scriptsize (1.40--1.50)}} & \makecell{1.29 \\{\scriptsize (1.26--1.36)}} \\
 & 100 & \makecell{43.1 \\{\scriptsize (40.2--45.8)}} & \makecell{\textbf{72.1} \\{\scriptsize (71.1--73.3)}} & \makecell{23.2 \\{\scriptsize (21.5--25.0)}} & \makecell{\underline{45.9} \\{\scriptsize (40.9--50.2)}} & \makecell{6.62 \\{\scriptsize (4.24--9.54)}} & \makecell{21.0 \\{\scriptsize (16.0--26.5)}} & \makecell{1.61 \\{\scriptsize (1.59--1.66)}} & \makecell{\textbf{1.65} \\{\scriptsize (1.62--1.71)}} & \makecell{1.53 \\{\scriptsize (1.50--1.58)}} & \makecell{\underline{1.63} \\{\scriptsize (1.60--1.68)}} & \makecell{1.24 \\{\scriptsize (1.21--1.31)}} & \makecell{1.43 \\{\scriptsize (1.40--1.50)}} & \makecell{1.29 \\{\scriptsize (1.26--1.36)}} \\
\midrule
\multirow[c]{5}{*}{\centering \makecell{Chest \\ 3,944}} & 1 & \makecell{145 \\{\scriptsize (111--197)}} & \makecell{\textbf{219} \\{\scriptsize (175--257)}} & \makecell{42.0 \\{\scriptsize (26.3--64.8)}} & \makecell{23.0 \\{\scriptsize (14.0--36.1)}} & \makecell{22.0 \\{\scriptsize (10.9--30.4)}} & \makecell{\underline{184} \\{\scriptsize (111--231)}} & \makecell{\underline{1.33} \\{\scriptsize (1.28--1.38)}} & \makecell{\textbf{1.37} \\{\scriptsize (1.30--1.41)}} & \makecell{1.31 \\{\scriptsize (1.24--1.34)}} & \makecell{1.27 \\{\scriptsize (1.23--1.32)}} & \makecell{1.18 \\{\scriptsize (1.12--1.23)}} & \makecell{1.29 \\{\scriptsize (1.22--1.34)}} & \makecell{1.19 \\{\scriptsize (1.11--1.23)}} \\
 & 5 & \makecell{85.2 \\{\scriptsize (67.9--113)}} & \makecell{\textbf{134} \\{\scriptsize (118--145)}} & \makecell{31.2 \\{\scriptsize (23.3--42.4)}} & \makecell{19.0 \\{\scriptsize (14.2--26.1)}} & \makecell{12.8 \\{\scriptsize (7.35--16.7)}} & \makecell{\underline{92.0} \\{\scriptsize (58.4--114)}} & \makecell{\underline{1.22} \\{\scriptsize (1.18--1.25)}} & \makecell{\textbf{1.24} \\{\scriptsize (1.19--1.26)}} & \makecell{1.20 \\{\scriptsize (1.16--1.22)}} & \makecell{1.18 \\{\scriptsize (1.14--1.20)}} & \makecell{1.11 \\{\scriptsize (1.06--1.15)}} & \makecell{1.19 \\{\scriptsize (1.14--1.22)}} & \makecell{1.12 \\{\scriptsize (1.07--1.14)}} \\
 & 10 & \makecell{\underline{64.3} \\{\scriptsize (51.9--84.0)}} & \makecell{\textbf{95.2} \\{\scriptsize (86.2--101)}} & \makecell{26.9 \\{\scriptsize (20.5--36.0)}} & \makecell{15.3 \\{\scriptsize (11.9--20.0)}} & \makecell{11.6 \\{\scriptsize (7.04--15.0)}} & \makecell{63.3 \\{\scriptsize (41.1--78.0)}} & \makecell{\underline{1.23} \\{\scriptsize (1.20--1.26)}} & \makecell{\textbf{1.24} \\{\scriptsize (1.21--1.28)}} & \makecell{1.20 \\{\scriptsize (1.17--1.23)}} & \makecell{1.19 \\{\scriptsize (1.16--1.23)}} & \makecell{1.11 \\{\scriptsize (1.08--1.16)}} & \makecell{1.19 \\{\scriptsize (1.15--1.23)}} & \makecell{1.12 \\{\scriptsize (1.08--1.16)}} \\
 & 50 & \makecell{\underline{28.6} \\{\scriptsize (24.4--35.4)}} & \makecell{\textbf{38.2} \\{\scriptsize (36.2--39.6)}} & \makecell{16.2 \\{\scriptsize (12.9--20.6)}} & \makecell{9.10 \\{\scriptsize (7.79--11.1)}} & \makecell{8.08 \\{\scriptsize (5.50--9.92)}} & \makecell{25.7 \\{\scriptsize (18.8--30.0)}} & \makecell{\underline{1.25} \\{\scriptsize (1.23--1.30)}} & \makecell{\textbf{1.27} \\{\scriptsize (1.24--1.32)}} & \makecell{1.22 \\{\scriptsize (1.19--1.26)}} & \makecell{1.23 \\{\scriptsize (1.19--1.27)}} & \makecell{1.12 \\{\scriptsize (1.07--1.17)}} & \makecell{1.18 \\{\scriptsize (1.14--1.24)}} & \makecell{1.12 \\{\scriptsize (1.08--1.17)}} \\
 & 100 & \makecell{\underline{18.8} \\{\scriptsize (16.2--22.9)}} & \makecell{\textbf{24.4} \\{\scriptsize (23.6--25.0)}} & \makecell{12.5 \\{\scriptsize (10.2--15.7)}} & \makecell{6.71 \\{\scriptsize (5.86--7.90)}} & \makecell{6.58 \\{\scriptsize (4.65--7.93)}} & \makecell{16.8 \\{\scriptsize (13.3--19.1)}} & \makecell{\underline{1.26} \\{\scriptsize (1.23--1.30)}} & \makecell{\textbf{1.27} \\{\scriptsize (1.23--1.33)}} & \makecell{1.22 \\{\scriptsize (1.19--1.26)}} & \makecell{1.23 \\{\scriptsize (1.19--1.28)}} & \makecell{1.12 \\{\scriptsize (1.07--1.18)}} & \makecell{1.17 \\{\scriptsize (1.12--1.23)}} & \makecell{1.12 \\{\scriptsize (1.08--1.17)}} \\
\midrule
\multirow[c]{5}{*}{\centering \makecell{AP \\ 8,731}} & 1 & \makecell{270 \\{\scriptsize (241--305)}} & \makecell{\underline{604} \\{\scriptsize (560--651)}} & \makecell{53.0 \\{\scriptsize (39.8--67.0)}} & \makecell{\textbf{733} \\{\scriptsize (685--782)}} & \makecell{6.00 \\{\scriptsize (2.00--11.0)}} & \makecell{29.0 \\{\scriptsize (19.0--39.9)}} & \makecell{1.75 \\{\scriptsize (1.68--1.79)}} & \makecell{\textbf{1.82} \\{\scriptsize (1.75--1.85)}} & \makecell{1.66 \\{\scriptsize (1.59--1.69)}} & \makecell{\underline{1.81} \\{\scriptsize (1.73--1.83)}} & \makecell{1.26 \\{\scriptsize (1.20--1.30)}} & \makecell{1.55 \\{\scriptsize (1.48--1.58)}} & \makecell{1.30 \\{\scriptsize (1.24--1.35)}} \\
 & 5 & \makecell{157 \\{\scriptsize (148--168)}} & \makecell{\underline{357} \\{\scriptsize (342--371)}} & \makecell{44.8 \\{\scriptsize (39.0--50.5)}} & \makecell{\textbf{357} \\{\scriptsize (343--372)}} & \makecell{4.40 \\{\scriptsize (2.60--6.25)}} & \makecell{22.0 \\{\scriptsize (17.8--26.1)}} & \makecell{1.41 \\{\scriptsize (1.36--1.42)}} & \makecell{\textbf{1.44} \\{\scriptsize (1.39--1.45)}} & \makecell{1.35 \\{\scriptsize (1.29--1.35)}} & \makecell{\underline{1.43} \\{\scriptsize (1.38--1.44)}} & \makecell{1.14 \\{\scriptsize (1.07--1.13)}} & \makecell{1.29 \\{\scriptsize (1.23--1.29)}} & \makecell{1.17 \\{\scriptsize (1.10--1.17)}} \\
 & 10 & \makecell{122 \\{\scriptsize (116--129)}} & \makecell{\textbf{254} \\{\scriptsize (245--262)}} & \makecell{38.8 \\{\scriptsize (34.9--42.8)}} & \makecell{\underline{238} \\{\scriptsize (230--245)}} & \makecell{4.00 \\{\scriptsize (2.80--5.23)}} & \makecell{18.2 \\{\scriptsize (15.6--20.7)}} & \makecell{1.43 \\{\scriptsize (1.40--1.46)}} & \makecell{\textbf{1.46} \\{\scriptsize (1.43--1.49)}} & \makecell{1.36 \\{\scriptsize (1.33--1.38)}} & \makecell{\underline{1.45} \\{\scriptsize (1.42--1.48)}} & \makecell{1.13 \\{\scriptsize (1.11--1.17)}} & \makecell{1.29 \\{\scriptsize (1.26--1.32)}} & \makecell{1.17 \\{\scriptsize (1.14--1.20)}} \\
 & 50 & \makecell{56.6 \\{\scriptsize (54.8--58.2)}} & \makecell{\textbf{96.6} \\{\scriptsize (94.7--98.4)}} & \makecell{25.4 \\{\scriptsize (24.1--26.6)}} & \makecell{\underline{78.0} \\{\scriptsize (76.2--79.8)}} & \makecell{2.92 \\{\scriptsize (2.40--3.42)}} & \makecell{12.8 \\{\scriptsize (11.8--13.7)}} & \makecell{1.54 \\{\scriptsize (1.52--1.58)}} & \makecell{\textbf{1.57} \\{\scriptsize (1.55--1.62)}} & \makecell{1.45 \\{\scriptsize (1.44--1.50)}} & \makecell{\underline{1.56} \\{\scriptsize (1.54--1.60)}} & \makecell{1.16 \\{\scriptsize (1.15--1.21)}} & \makecell{1.35 \\{\scriptsize (1.34--1.40)}} & \makecell{1.20 \\{\scriptsize (1.19--1.25)}} \\
 & 100 & \makecell{36.9 \\{\scriptsize (36.0--37.8)}} & \makecell{\textbf{58.5} \\{\scriptsize (57.7--59.3)}} & \makecell{19.6 \\{\scriptsize (18.8--20.3)}} & \makecell{\underline{46.2} \\{\scriptsize (45.3--47.1)}} & \makecell{2.74 \\{\scriptsize (2.41--3.07)}} & \makecell{9.94 \\{\scriptsize (9.36--10.5)}} & \makecell{1.55 \\{\scriptsize (1.53--1.60)}} & \makecell{\textbf{1.59} \\{\scriptsize (1.57--1.63)}} & \makecell{1.46 \\{\scriptsize (1.44--1.51)}} & \makecell{\underline{1.57} \\{\scriptsize (1.55--1.62)}} & \makecell{1.16 \\{\scriptsize (1.15--1.21)}} & \makecell{1.35 \\{\scriptsize (1.33--1.40)}} & \makecell{1.19 \\{\scriptsize (1.18--1.24)}} \\
\midrule
\multirow[c]{5}{*}{\centering \makecell{CAP \\ 1,881}} & 1 & \makecell{\textbf{109} \\{\scriptsize (90.3--129)}} & \makecell{\underline{78.0} \\{\scriptsize (61.6--95.7)}} & \makecell{28.0 \\{\scriptsize (18.1--39.5)}} & \makecell{25.0 \\{\scriptsize (15.9--34.9)}} & \makecell{6.00 \\{\scriptsize (1.99--11.2)}} & \makecell{24.0 \\{\scriptsize (14.6--34.3)}} & \makecell{\textbf{1.48} \\{\scriptsize (1.38--1.54)}} & \makecell{\underline{1.46} \\{\scriptsize (1.35--1.52)}} & \makecell{1.44 \\{\scriptsize (1.34--1.50)}} & \makecell{1.36 \\{\scriptsize (1.29--1.45)}} & \makecell{1.23 \\{\scriptsize (1.13--1.30)}} & \makecell{1.35 \\{\scriptsize (1.25--1.41)}} & \makecell{1.21 \\{\scriptsize (1.12--1.29)}} \\
 & 5 & \makecell{\textbf{62.4} \\{\scriptsize (56.1--68.9)}} & \makecell{\underline{53.6} \\{\scriptsize (47.7--60.3)}} & \makecell{24.4 \\{\scriptsize (20.2--28.4)}} & \makecell{19.0 \\{\scriptsize (15.3--22.6)}} & \makecell{3.60 \\{\scriptsize (2.02--5.44)}} & \makecell{13.6 \\{\scriptsize (10.5--17.0)}} & \makecell{\textbf{1.29} \\{\scriptsize (1.22--1.32)}} & \makecell{\underline{1.27} \\{\scriptsize (1.21--1.31)}} & \makecell{1.26 \\{\scriptsize (1.19--1.29)}} & \makecell{1.23 \\{\scriptsize (1.16--1.26)}} & \makecell{1.13 \\{\scriptsize (1.06--1.17)}} & \makecell{1.21 \\{\scriptsize (1.14--1.24)}} & \makecell{1.13 \\{\scriptsize (1.06--1.17)}} \\
 & 10 & \makecell{\textbf{47.8} \\{\scriptsize (44.2--51.6)}} & \makecell{\underline{38.6} \\{\scriptsize (35.3--41.9)}} & \makecell{19.9 \\{\scriptsize (17.3--22.4)}} & \makecell{15.1 \\{\scriptsize (12.8--17.4)}} & \makecell{3.30 \\{\scriptsize (2.24--4.54)}} & \makecell{10.5 \\{\scriptsize (8.55--12.5)}} & \makecell{\textbf{1.30} \\{\scriptsize (1.25--1.35)}} & \makecell{\underline{1.27} \\{\scriptsize (1.23--1.33)}} & \makecell{1.27 \\{\scriptsize (1.22--1.32)}} & \makecell{1.24 \\{\scriptsize (1.19--1.29)}} & \makecell{1.13 \\{\scriptsize (1.08--1.19)}} & \makecell{1.20 \\{\scriptsize (1.16--1.26)}} & \makecell{1.13 \\{\scriptsize (1.08--1.19)}} \\
 & 50 & \makecell{\textbf{19.8} \\{\scriptsize (18.9--20.6)}} & \makecell{\underline{15.9} \\{\scriptsize (15.1--16.8)}} & \makecell{11.7 \\{\scriptsize (10.9--12.5)}} & \makecell{8.14 \\{\scriptsize (7.42--8.86)}} & \makecell{2.48 \\{\scriptsize (2.04--2.94)}} & \makecell{6.98 \\{\scriptsize (6.32--7.62)}} & \makecell{\textbf{1.34} \\{\scriptsize (1.30--1.41)}} & \makecell{\underline{1.32} \\{\scriptsize (1.28--1.38)}} & \makecell{1.30 \\{\scriptsize (1.26--1.37)}} & \makecell{1.28 \\{\scriptsize (1.24--1.35)}} & \makecell{1.13 \\{\scriptsize (1.09--1.20)}} & \makecell{1.20 \\{\scriptsize (1.16--1.27)}} & \makecell{1.14 \\{\scriptsize (1.10--1.21)}} \\
 & 100 & \makecell{\textbf{12.6} \\{\scriptsize (12.2--13.0)}} & \makecell{\underline{10.8} \\{\scriptsize (10.4--11.2)}} & \makecell{8.98 \\{\scriptsize (8.54--9.42)}} & \makecell{5.62 \\{\scriptsize (5.20--5.99)}} & \makecell{2.14 \\{\scriptsize (1.85--2.44)}} & \makecell{5.16 \\{\scriptsize (4.79--5.56)}} & \makecell{\textbf{1.33} \\{\scriptsize (1.28--1.39)}} & \makecell{\underline{1.31} \\{\scriptsize (1.26--1.37)}} & \makecell{1.29 \\{\scriptsize (1.24--1.36)}} & \makecell{1.27 \\{\scriptsize (1.23--1.34)}} & \makecell{1.12 \\{\scriptsize (1.07--1.19)}} & \makecell{1.18 \\{\scriptsize (1.13--1.25)}} & \makecell{1.13 \\{\scriptsize (1.08--1.20)}} \\
\midrule
\multirow[c]{5}{*}{\centering \makecell{Oncological \\ 3,264}} & 1 & \makecell{\textbf{224} \\{\scriptsize (195--253)}} & \makecell{\underline{151} \\{\scriptsize (128--174)}} & \makecell{53.0 \\{\scriptsize (39.8--66.9)}} & \makecell{38.0 \\{\scriptsize (26.1--51.2)}} & \makecell{6.00 \\{\scriptsize (1.99--11.0)}} & \makecell{41.0 \\{\scriptsize (29.1--53.9)}} & \makecell{\textbf{1.69} \\{\scriptsize (1.60--1.74)}} & \makecell{\underline{1.66} \\{\scriptsize (1.57--1.72)}} & \makecell{1.60 \\{\scriptsize (1.51--1.66)}} & \makecell{1.51 \\{\scriptsize (1.43--1.58)}} & \makecell{1.32 \\{\scriptsize (1.23--1.38)}} & \makecell{1.40 \\{\scriptsize (1.32--1.48)}} & \makecell{1.32 \\{\scriptsize (1.23--1.39)}} \\
 & 5 & \makecell{\textbf{115} \\{\scriptsize (106--125)}} & \makecell{\underline{93.8} \\{\scriptsize (86.0--102)}} & \makecell{44.6 \\{\scriptsize (38.8--50.2)}} & \makecell{27.6 \\{\scriptsize (23.1--32.2)}} & \makecell{6.20 \\{\scriptsize (4.19--8.43)}} & \makecell{22.8 \\{\scriptsize (18.7--27.1)}} & \makecell{\textbf{1.38} \\{\scriptsize (1.32--1.41)}} & \makecell{\underline{1.36} \\{\scriptsize (1.31--1.39)}} & \makecell{1.33 \\{\scriptsize (1.27--1.36)}} & \makecell{1.27 \\{\scriptsize (1.22--1.30)}} & \makecell{1.16 \\{\scriptsize (1.09--1.18)}} & \makecell{1.21 \\{\scriptsize (1.15--1.24)}} & \makecell{1.17 \\{\scriptsize (1.10--1.20)}} \\
 & 10 & \makecell{\textbf{83.4} \\{\scriptsize (78.4--88.6)}} & \makecell{\underline{71.7} \\{\scriptsize (67.3--76.6)}} & \makecell{38.3 \\{\scriptsize (34.7--42.0)}} & \makecell{21.5 \\{\scriptsize (18.8--24.2)}} & \makecell{4.60 \\{\scriptsize (3.32--5.90)}} & \makecell{16.9 \\{\scriptsize (14.4--19.5)}} & \makecell{\textbf{1.41} \\{\scriptsize (1.37--1.45)}} & \makecell{\underline{1.39} \\{\scriptsize (1.35--1.44)}} & \makecell{1.35 \\{\scriptsize (1.31--1.40)}} & \makecell{1.30 \\{\scriptsize (1.26--1.35)}} & \makecell{1.17 \\{\scriptsize (1.12--1.22)}} & \makecell{1.22 \\{\scriptsize (1.18--1.27)}} & \makecell{1.18 \\{\scriptsize (1.14--1.23)}} \\
 & 50 & \makecell{\textbf{33.5} \\{\scriptsize (32.4--34.6)}} & \makecell{\underline{28.7} \\{\scriptsize (27.6--29.8)}} & \makecell{21.8 \\{\scriptsize (20.8--22.9)}} & \makecell{11.2 \\{\scriptsize (10.3--12.0)}} & \makecell{3.28 \\{\scriptsize (2.79--3.78)}} & \makecell{9.94 \\{\scriptsize (9.16--10.7)}} & \makecell{\textbf{1.48} \\{\scriptsize (1.45--1.55)}} & \makecell{\underline{1.46} \\{\scriptsize (1.42--1.52)}} & \makecell{1.42 \\{\scriptsize (1.39--1.48)}} & \makecell{1.37 \\{\scriptsize (1.34--1.44)}} & \makecell{1.18 \\{\scriptsize (1.15--1.25)}} & \makecell{1.23 \\{\scriptsize (1.20--1.30)}} & \makecell{1.20 \\{\scriptsize (1.18--1.27)}} \\
 & 100 & \makecell{\textbf{21.0} \\{\scriptsize (20.5--21.5)}} & \makecell{\underline{18.2} \\{\scriptsize (17.7--18.7)}} & \makecell{15.7 \\{\scriptsize (15.1--16.3)}} & \makecell{8.09 \\{\scriptsize (7.61--8.60)}} & \makecell{2.90 \\{\scriptsize (2.56--3.23)}} & \makecell{7.38 \\{\scriptsize (6.93--7.88)}} & \makecell{\textbf{1.49} \\{\scriptsize (1.45--1.55)}} & \makecell{\underline{1.46} \\{\scriptsize (1.42--1.52)}} & \makecell{1.42 \\{\scriptsize (1.38--1.48)}} & \makecell{1.38 \\{\scriptsize (1.34--1.45)}} & \makecell{1.18 \\{\scriptsize (1.13--1.24)}} & \makecell{1.21 \\{\scriptsize (1.17--1.28)}} & \makecell{1.20 \\{\scriptsize (1.16--1.26)}} \\
\midrule
\multirow[c]{5}{*}{\centering \makecell{Non-oncologic \\ 7,530}} & 1 & \makecell{190 \\{\scriptsize (160--224)}} & \makecell{\underline{659} \\{\scriptsize (609--711)}} & \makecell{44.0 \\{\scriptsize (32.3--56.7)}} & \makecell{\textbf{745} \\{\scriptsize (619--880)}} & \makecell{22.0 \\{\scriptsize (7.78--36.1)}} & \makecell{191 \\{\scriptsize (93.7--288)}} & \makecell{1.69 \\{\scriptsize (1.61--1.74)}} & \makecell{\underline{1.78} \\{\scriptsize (1.69--1.83)}} & \makecell{1.60 \\{\scriptsize (1.51--1.64)}} & \makecell{\textbf{1.80} \\{\scriptsize (1.72--1.83)}} & \makecell{1.27 \\{\scriptsize (1.17--1.35)}} & \makecell{1.60 \\{\scriptsize (1.51--1.65)}} & \makecell{1.33 \\{\scriptsize (1.23--1.40)}} \\
 & 5 & \makecell{116 \\{\scriptsize (102--132)}} & \makecell{\textbf{377} \\{\scriptsize (361--395)}} & \makecell{34.2 \\{\scriptsize (29.0--40.4)}} & \makecell{\underline{365} \\{\scriptsize (308--423)}} & \makecell{14.4 \\{\scriptsize (6.30--22.3)}} & \makecell{99.0 \\{\scriptsize (53.6--142)}} & \makecell{1.37 \\{\scriptsize (1.31--1.38)}} & \makecell{\underline{1.42} \\{\scriptsize (1.35--1.44)}} & \makecell{1.30 \\{\scriptsize (1.24--1.32)}} & \makecell{\textbf{1.43} \\{\scriptsize (1.37--1.44)}} & \makecell{1.14 \\{\scriptsize (1.05--1.16)}} & \makecell{1.31 \\{\scriptsize (1.24--1.33)}} & \makecell{1.17 \\{\scriptsize (1.09--1.19)}} \\
 & 10 & \makecell{90.3 \\{\scriptsize (80.5--101)}} & \makecell{\textbf{265} \\{\scriptsize (255--276)}} & \makecell{26.8 \\{\scriptsize (23.1--31.4)}} & \makecell{\underline{236} \\{\scriptsize (199--272)}} & \makecell{11.9 \\{\scriptsize (5.55--18.2)}} & \makecell{69.3 \\{\scriptsize (38.8--98.1)}} & \makecell{1.42 \\{\scriptsize (1.38--1.46)}} & \makecell{\underline{1.47} \\{\scriptsize (1.42--1.51)}} & \makecell{1.35 \\{\scriptsize (1.30--1.39)}} & \makecell{\textbf{1.48} \\{\scriptsize (1.44--1.51)}} & \makecell{1.16 \\{\scriptsize (1.11--1.22)}} & \makecell{1.35 \\{\scriptsize (1.30--1.40)}} & \makecell{1.20 \\{\scriptsize (1.15--1.25)}} \\
 & 50 & \makecell{43.0 \\{\scriptsize (39.7--46.7)}} & \makecell{\textbf{95.8} \\{\scriptsize (93.0--98.5)}} & \makecell{17.5 \\{\scriptsize (16.0--19.1)}} & \makecell{\underline{74.2} \\{\scriptsize (64.0--84.8)}} & \makecell{8.24 \\{\scriptsize (4.10--12.4)}} & \makecell{27.6 \\{\scriptsize (18.7--36.5)}} & \makecell{1.52 \\{\scriptsize (1.48--1.57)}} & \makecell{\underline{1.57} \\{\scriptsize (1.52--1.63)}} & \makecell{1.43 \\{\scriptsize (1.39--1.49)}} & \makecell{\textbf{1.58} \\{\scriptsize (1.54--1.63)}} & \makecell{1.18 \\{\scriptsize (1.14--1.26)}} & \makecell{1.41 \\{\scriptsize (1.36--1.48)}} & \makecell{1.22 \\{\scriptsize (1.18--1.31)}} \\
 & 100 & \makecell{28.7 \\{\scriptsize (27.0--30.6)}} & \makecell{\textbf{56.6} \\{\scriptsize (55.3--58.0)}} & \makecell{13.2 \\{\scriptsize (12.3--14.1)}} & \makecell{\underline{42.8} \\{\scriptsize (37.2--48.5)}} & \makecell{6.41 \\{\scriptsize (3.27--9.48)}} & \makecell{18.7 \\{\scriptsize (13.6--23.5)}} & \makecell{1.53 \\{\scriptsize (1.49--1.59)}} & \makecell{\underline{1.58} \\{\scriptsize (1.53--1.65)}} & \makecell{1.44 \\{\scriptsize (1.40--1.51)}} & \makecell{\textbf{1.59} \\{\scriptsize (1.55--1.64)}} & \makecell{1.18 \\{\scriptsize (1.13--1.27)}} & \makecell{1.41 \\{\scriptsize (1.36--1.49)}} & \makecell{1.23 \\{\scriptsize (1.18--1.31)}} \\
\midrule
\end{tabular}
\end{adjustbox}
\caption{Zero-shot text-image (Recall@K) and image-image (MAP@K) retrieval performance of CAPTOR, CAPTOR-100K, CAPTOR-NoPre and four state-of-the-art models, expressed as fold-increase over the random baseline (best in bold and second best underlined), by anatomical and oncologic subset. K is the number of retrieved images. SPECTRE~\cite{claessens_scaling_2025} could not be evaluated (unavailable projection-layer weights), and CT-FM~\cite{pai_vision_2025}, a vision-only model, was excluded from text-image retrieval.}
\label{retrieval_captor100K}
\end{table*}

\begin{figure*}[t]
    \centering
    \includegraphics[width=\linewidth]{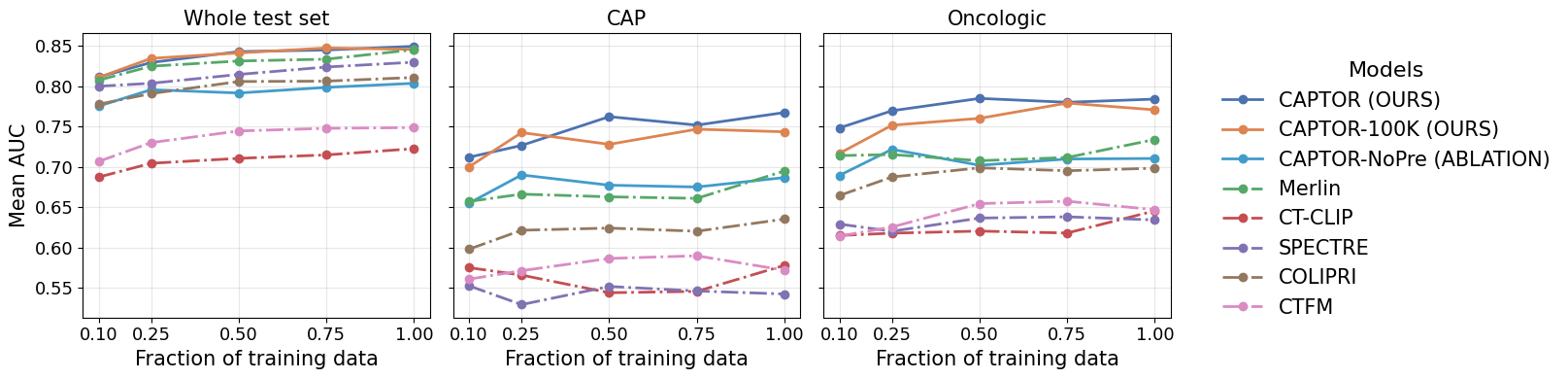}
    \caption{Classification performance (macro-AUC) as a function of XGBoost training-set fraction (10-100\%), reported for CAPTOR, CAPTOR-100K, CAPTOR-NoPre (i.e. model trained on the same dataset as CAPTOR but without our anatomical crops preprocessing), and five state-of-the-art foundation models, on the whole test set (left), the CAP subset (middle), and the oncologic subset (right).}
    \label{fig:ablation}
\end{figure*} 

\subsection{External datasets with original labels}\label{app:results_external_datasets}
\begin{figure*}[t]
    \centering
    \includegraphics[width=\linewidth]{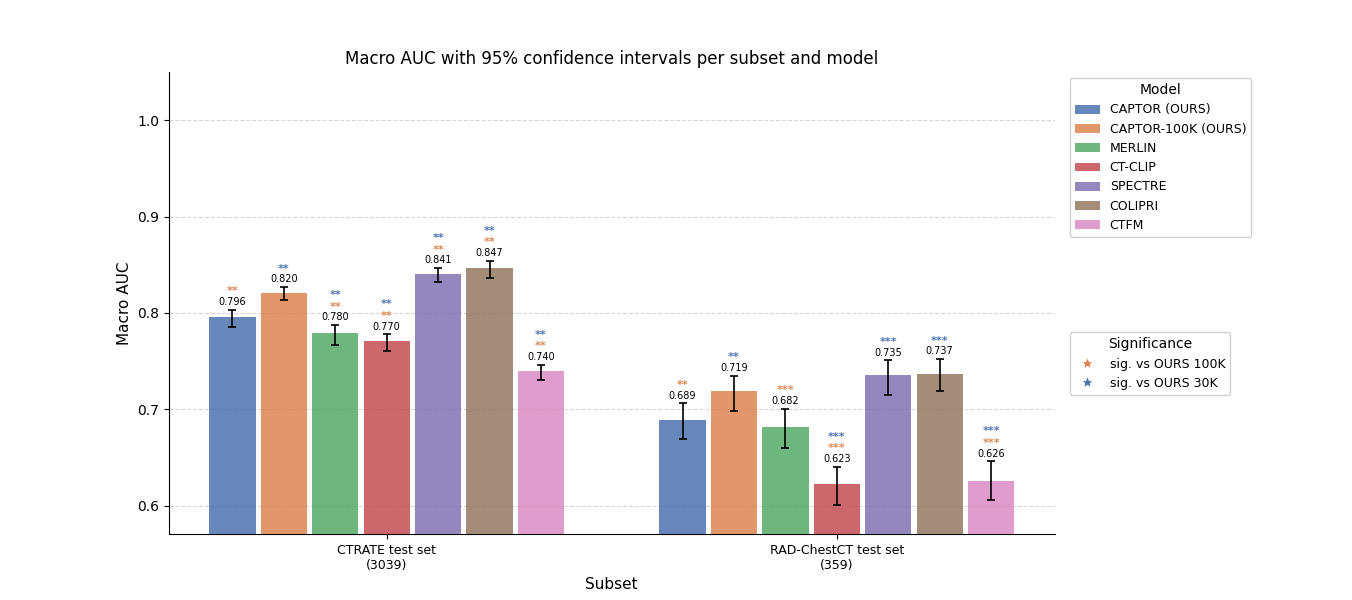}
    \caption{Classification macro-AUC of CAPTOR, CAPTOR-100K, and five baseline models for each abnormality label on the external, non-oncologic CT-RATE~\cite{hamamci_generalist_2026} and RAD-ChestCT~\cite{draelos_rad-chestct_2020} datasets, using each dataset's original labels. One XGBoost was trained for each medical condition and model using the image embedding extracted with the model’s image encoder. For each subset and model, the macro-AUC is the unweighted mean of the AUC scores for each medical condition relevant to the subset among the 26 conditions we studied. The confidence intervals and the pairwise comparison were obtained with a patient-level bootstrap and a patient-level permutation test respectively, both of size 2,000.  *** = p < 0.001, ** = p < 0.01, * = p < 0.05}
    \label{fig:bar_plot_external}
\end{figure*}

CAPTOR, more specialized for CAP oncologic imaging, is significantly outperformed (p<0.01) by our broader 100K model on both CT-RATE~\cite{hamamci_generalist_2026} and RAD-ChestCT~\cite{draelos_rad-chestct_2020} (Figure \ref{fig:bar_plot_external})

Both our models are outperformed by COLIPRI~\cite{wald_comprehensive_2026} and SPECTRE~\cite{claessens_scaling_2025}but not significantly for CAPTOR-100K on RAD-ChestCT.

\end{document}